\documentclass[twoside,leqno,twocolumn]{article}

\usepackage[letterpaper]{geometry}

\usepackage{ltexpprt}
\usepackage{graphicx}
\usepackage{multirow}
\usepackage{adjustbox}
\usepackage{compact}
\usepackage[colorlinks,citecolor=blue]{hyperref}
\usepackage{amssymb,amsmath}

\begin{document}

\newcommand\relatedversion{}
\renewcommand\relatedversion{\thanks{The full version of the paper can be accessed at \protect\url{https://arxiv.org/abs/1902.09310}}} % Replace URL with link to full paper or comment out this line

\title{Interpretable Molecular Graph Generation \mbox{via Monotonic Constraints}}
\author{Yuanqi Du\thanks{Yuanqi Du and Xiaojie Guo contributed equally to this work.}\textsuperscript{ ~}\thanks{George Mason University, VA 22030, USA.}
\and Xiaojie Guo\footnotemark[1]\textsuperscript{ ~}\footnotemark[2]
\and Amarda Shehu\footnotemark[2]
\and Liang Zhao\thanks{Emory University, GA 30322, USA. Email: liang.zhao@emory.edu}}
\date{}

\maketitle

% Copyright Statement
% When submitting your final paper to a SIAM proceedings, it is requested that you include
% the appropriate copyright in the footer of the paper.  The copyright added should be
% consistent with the copyright selected on the copyright form submitted with the paper.
% Please note that "20XX" should be changed to the year of the meeting.

% Default Copyright Statement
\fancyfoot[R]{\scriptsize{Copyright \textcopyright\ 2022 by SIAM\\
Unauthorized reproduction of this article is prohibited}}

% Depending on which copyright you agree to when you sign the copyright form, the copyright
% can be changed to one of the following after commenting out the default copyright statement
% above.

%\fancyfoot[R]{\scriptsize{Copyright \textcopyright\ 20XX\\
%Copyright for this paper is retained by authors}}

%\fancyfoot[R]{\scriptsize{Copyright \textcopyright\ 20XX\\
%Copyright retained by principal author's organization}}

%\pagenumbering{arabic}
%\setcounter{page}{1}%Leave this line commented out.

\begin{abstract}
Designing molecules with specific properties is a long-lasting research problem and is central to advancing crucial domains such as drug discovery and material science. Recent advances in deep graph generative models treat molecule design as graph generation problems which provide new opportunities toward the breakthrough of this long-lasting problem. Existing models, however, have many shortcomings, including poor interpretability and controllability toward desired molecular properties. This paper focuses on new methodologies for molecule generation with interpretable and controllable deep generative models, by proposing new monotonically-regularized graph variational autoencoders. The proposed models learn to represent the molecules with latent variables and then learn the correspondence between them and molecule properties parameterized by polynomial functions. To further improve the intepretability and controllability of molecule generation towards desired properties, we derive new objectives which further enforce monotonicity of the relation between some latent variables and target molecule properties such as toxicity and clogP. Extensive experimental evaluation demonstrates the superiority of the proposed framework on accuracy, novelty, disentanglement, and control towards desired molecular properties. The code is anonymized at \url{https://anonymous.4open.science/r/MDVAE-FD2C}.
\end{abstract}

% The framework generates up to $100$\% syntactically-valid and novel molecules and achieves better disentanglement and control of desired properties of generated molecules than state-of-the-art methods. To be specific, the proposed models have outperformed the state-of-the-art methods by generating more accurate and better molecules by up to 68\% improvement in learning the molecular property distribution, up to 43\% improvement in interpretability, and 34\% improvement in controlling the molecular properties.

\section{Introduction}

% \begin{figure*}[htbp]
%   \centering
%   \begin{tabular}{@{}c@{}}
%     \includegraphics[width=0.4\linewidth]{Figures/Conditional_Age.png}
%     \includegraphics[width=0.4\linewidth]{Figures/CtlConditional_Pose.png}\\[-2mm]
%     \small (a) Image generation without control
%     \hspace{3.25cm} 
%     \small (b) Image generation with control\\
%   \end{tabular}

%   \vspace{\floatsep}

%   \begin{tabular}{@{}c@{}}
%     \includegraphics[width=0.4\linewidth]{Figures/Conditional_Mol.png}
%     \includegraphics[width=0.4\linewidth]{Figures/CtlConditional_Mol.png}
%     \\[-1mm]
%     \small (c) Molecule generation without control
%     \hspace{3cm}
%     \small (d) Molecule generation with control\\
%   \end{tabular}

%   \caption{(a) Image generation without control: Age control is difficult, since the latent variable is not monotonically correlated with it. (b) Image generation with control: Pose control is now easy, since the latent variable is monotonically correlated with it. (c) Molecule generation without control: Drug-likeness control is difficult, since the value of latent variable is not monotonically correlated with it. (d) Molecule generation with control: Drug-likeness control is now easy, since the latent variable is monotonically correlated with it.}
%   \label{fig:motivation}
% \end{figure*}

\begin{figure*}[htbp]
  \centering
  \begin{tabular}{cc}
    \includegraphics[width=0.48\textwidth]{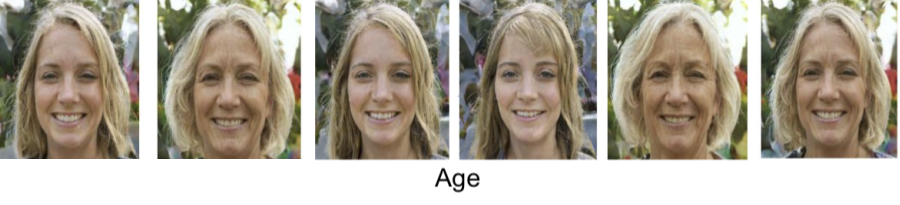} &
    \includegraphics[width=0.48\textwidth]{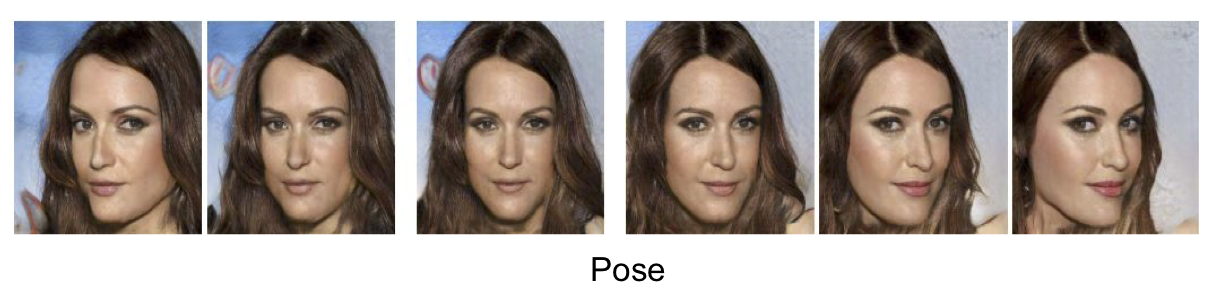}\\
    \small (a) Image generation without control &
    \small (b) Image generation with control\\
    \includegraphics[width=0.48\textwidth]{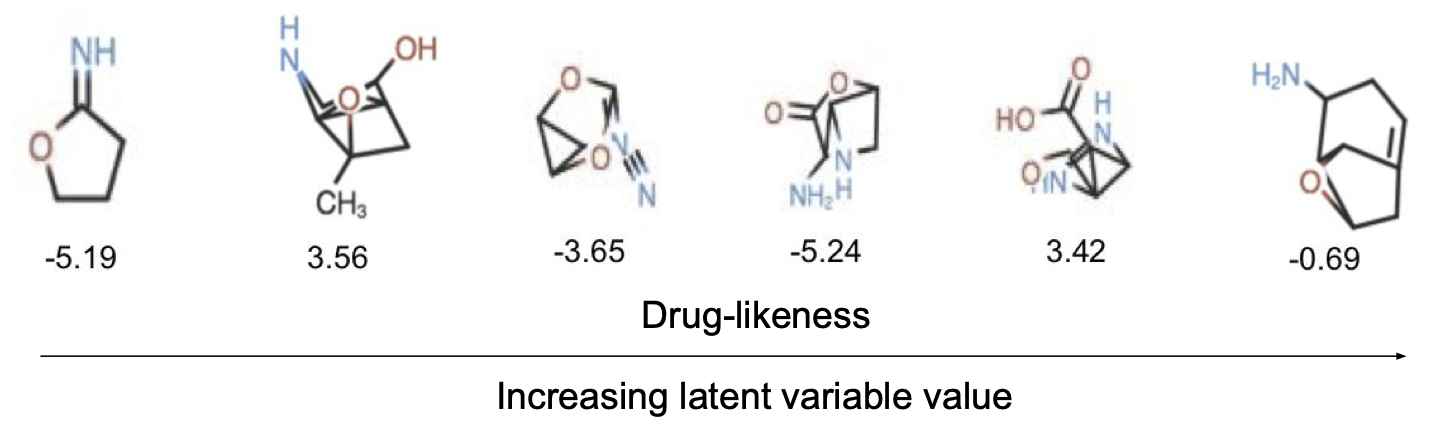} &
    \includegraphics[width=0.48\textwidth]{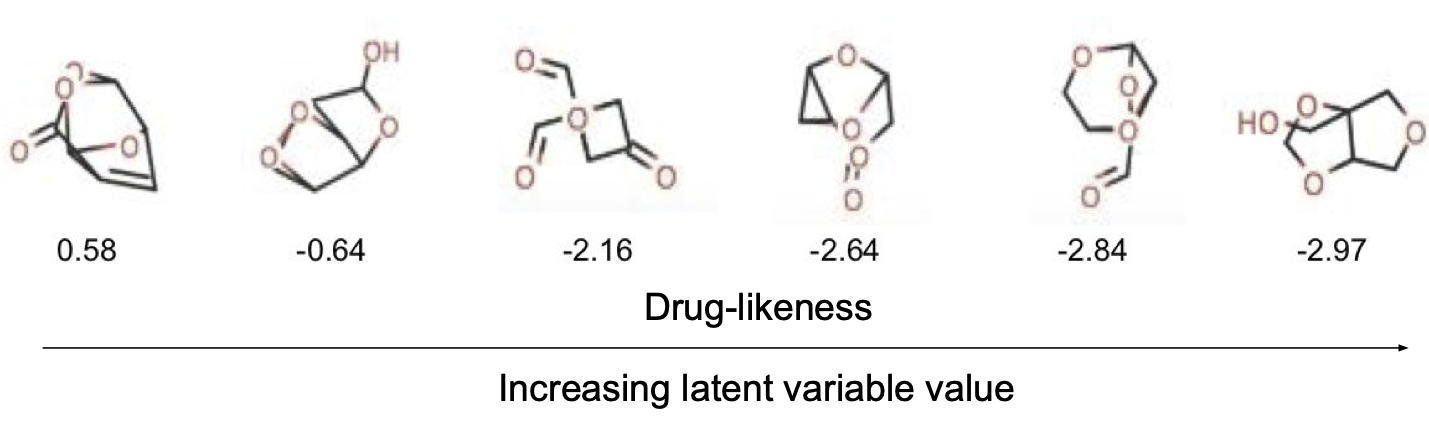}\\
    \small (c) Molecule generation without control &
    \small (d) Molecule generation with control\\
  \end{tabular}
  \caption{(a) Image generation without control: Age control is difficult, since the latent variable is not monotonically correlated with it. (b) Image generation with control: Pose control is now easy, since the latent variable is monotonically correlated with it. (c) Molecule generation without control: Drug-likeness control is difficult, since the value of latent variable is not monotonically correlated with it. (d) Molecule generation with control: Drug-likeness control is now easy, since the latent variable is monotonically correlated with it.}
  \label{fig:motivation}
\end{figure*}

Designing molecules with specific structural and functional properties is central to advancing drug discovery and material science~\cite{Whitesides15}. Decades of research in medicinal chemistry shows that finding novel drugs remains an outstanding challenge~\cite{Schneider16}, since the search space is vast and highly rugged; small perturbations in the chemical structure may result in great changes in desired properties. 
%While high-throughput technologies have improved significantly, the space is too large to address the de-novo design of molecules exclusively in the wet laboratory. Computational approaches provide a promising complementary approach.

While for many years computational screening  was primarily dominated by similarity search~\cite{Stumpfe11}, recent advances in deep generative models are showing promise in tackling de-novo molecule design. The first effort addressed the problem as a string generation task
%These works leverage the "molecular-input line-entry system" (SMILES) representation, which is a linear string representation of molecules and their active structures~\cite{SMILES}. SMILES is a formal grammar that describes molecules with an alphabet of characters; for instance, 'c' and 'C' denote aromatic and aliphatic carbon atoms, 'O' denotes the oxygen atom, '−' denotes single bonds, '=' denotes double bonds, and so on. 
by utilizing the SMILES representation~\cite{gomez2018automatic,kusner2017grammar}. However, SMILES is not designed to capture molecular similarity and prevents generative models (e.g., variational auto-encoders (VAE)) from learning smooth molecular embeddings. More importantly, essential chemical properties, such as molecular weight, cannot be expressed and preserved by the SMILES representation.

Recent advances in deep generative models on graphs have opened a new research direction for de-novo molecular design. Specifically, these models leverage more expressive representations of molecules via the concept of graphs, which is a natural formulation of molecule where atoms are connected by bonds. Graph-generative models hold much promise in generating credible molecules~\cite{simonovsky2018graphvae,jin2018junction,popova2019molecularrnn}. 
%It is worth noting that the latter is captured rigorously, by subjecting a generated molecule to the sanitization checks in RDKit~\cite{Landrum2016RDKit2016_09_4}.
The state-of-the-art deep generative models for molecule generation consist of two complementary subtasks: (1) the encoding, which refers to learning to represent molecules in a continuous manner that facilitates the preservation or optimization of their properties; (2) the decoding, which refers to learning to map an optimized continuous representation back into a reconstructed or novel molecule. 

Despite promising results, the existing models have several limitations:  
(1) \textbf{The molecule generation process is obscure}. Learning the correspondence between a molecule's structural patterns and its functional properties is one of the core issues in molecular modeling and design. However, although existing deep generative models for graphs can map the graph structural information into continuous representations, they are latent variables with no real-world meaning. Moreover, the latent variables may have mutual correlations with each other which further prevents us from understanding their meanings. Models that can characterize the correspondence between the molecule structure, latent variables, and molecule properties with better transparency are imperative and have not been well explored.
(2) \textbf{Difficulty in controlling the properties of the generated molecule graphs}. It is important to generate molecules with desired biophysical and biochemical properties, such as toxicity, mass, and clogP \cite{leeson2007influence}; However, this is a very challenging and promising domain that has not been well explored historically. The few existing works typically formulate this as a conditional graph generation problem where the targeted properties are treated as conditions. However, it is difficult to assume the distributions of the properties as most the distribution of the real-world properties are unknown or too sophisticated to be predefined. Moreover, existing works need to assume the independency among the properties which usually is also not true, for example, cLogP and cLogS are highly correlated. Thus, more powerful methods that can automatically estimate the distributions of the inter-correlated properties are imperative. Moreover, the correspondence between latent variables and properties  learned by the models need to be simple (e.g., monotonic and smooth) and hence easily controllable. For example, if the correspondence is a monotonic mapping then we can enlarge the latent variable's value to increase (or descrease) the value of a property. As shown in Fig. \ref{fig:motivation}(a), tuning the age in the generated image is not easy since increasing z may or may not increase or decrease the age. Similar trouble is also in Fig. \ref{fig:motivation}(c). But in Fig. \ref{fig:motivation}(b) and (d), it is much easier to tune the properties thanks to the monotonic relation between latent variables and targeted properties.

In this paper, we address the above limitations by proposing a \underline{M}onotonic \underline{D}isentangled \underline{VAE}, (MDVAE), which is a new framework that enhances the interpretability and controllability of deep graph generation of molecules. Specifically, a disentanglement loss is first introduced to enforce the disentanglement of latent variables for capturing more interpretable, factorized latent variables. In order to generate molecules with the desired properties, we then enforce a monotonic constraint over the correspondence between some latent variables and the targeted properties. Multiple strategies have also been proposed to instantiate the correspondence including linear and polynomial for the trade-off between model controllability and expressiveness. The contributions of this work are summarized as follows:
\begin{itemize}
    \item \textbf{A new framework of monotonically-constrained graph VAE is proposed for controllable generation.} The proposed model encodes the molecule structure into the latent disentangled variables, which can be used to reversely generate the molecules with desired properties with potential inter-correlations.
    \item \textbf{A polynomial parametrization for mapping latent variables to properties is introduced.} Our proposed polynomial parametrization explicitly enables the model to learn the linear and non-linear relationship between the latent variables and the desired properties with better trade-off between model capacity and transparency.
    \item \textbf{Various monotonic constraint strategies are proposed for regularizing the mapping between latent variables and molecule properties toward better controllability.} Gradient-based and direction-based monotonic constraints are both proposed to regularize the mapping between latent variables and molecule properties. Such constraints have further been generalized to handle the situation when molecule properties are correlated.
    \item \textbf{Extensive experiments demonstrated the effectiveness, interpretability, and controllability of our proposed models.} Qualitative and quantitative evaluations on multiple benchmark datasets demonstrated that the proposed models have outperformed the state-of-the-art methods by generating more accurate and better molecules by up to 68\% improvement in learning a more accurate molecular property distribution, up to 43\% improvement in interpretability, and up to 34\% improvement in controlling the molecular properties.
\end{itemize}
%First, the learnt latent representations of molecules are disentangled and interpretable. Second, the framework allows incorporating monotonic relations between the latent variables and desired properties. Third, the framework accommodates variable-size molecules. Finally, extensive experiments are carried out, demonstrating advantages over state-of-the-art methods.

%This paper is organized as follows. First, we provide a brief summary of related works in deep generative models for the problem of molecule generation in Section~\ref{sec:RelatedWork}. The framework is then described in detail in Section~\ref{sec:Methods}. Evaluation is presented in Section~\ref{sec:Results}. The paper concludes with a summary and future research directions in Section~\ref{sec:conclusion}.
\section{Related Work}
\label{sec:RelatedWork}

Early deep learning-based works in~\cite{gomez2018automatic, segler2018generating,krenn2020self} built generative models of SMILES strings with recurrent decoders. SMILES is a formal grammar that describes molecules with an alphabet of characters. For instance, `c' and `C' denote aromatic and aliphatic carbon atoms; `O' denotes the oxygen atom; `-' denotes single bonds; `=' denotes double bonds, and so on. Since initial models could generate invalid molecules, later works~\cite{kusner2017grammar,dai2018syntax} introduced syntactic and semantic constraints by context-free and attribute grammars; yet, the resulting models could not fully capture chemical validity. Other methods aimed to generate valid molecules by leveraging active learning~\cite{janz2017actively} and reinforcement learning~\cite{guimaraes2017objective}. 

Graph-generative models now present an alternative approach to molecule generation. For example, work in~\cite{simonovsky2018graphvae} generates molecular graphs by predicting their adjacency matrices. Work in~\cite{liu2018constrained} generates molecules through a constrained graph generative model that enforces validity by generating the molecule atom by atom. The majority of existing models are based on the VAE framework~\cite{simonovsky2018graphvae,samanta2018designing,guo2020property,du2020interpretable,ydu2021deeplatentcontrol,guo2021generating,guo2021deep,dudeeplatent2021,ydu2022spatiotemporal} or generative adversarial networks (GANs)~\cite{bojchevski2018netgan,guo2018deep,rahman2021generative}, and others~\cite{you2018graphrnn, liu2018constrained, fu2020core,fu2021mimosa,fu2021differentiable, du2021graphgt}. For instance, GraphRNN~\cite{you2018graphrnn}
builds an autoregressive generative model based on a generative recurrent neural network (RNN) by representing the graph as a sequence and generating nodes one by one. In contrast, GraphVAE~\cite{simonovsky2018graphvae} represents each graph in terms of its adjacent matrix and feature vectors of nodes. A VAE model is then utilized to learn the distribution of the graphs conditioned on a latent representation at the graph level. Other works~\cite{pmlr-v97-grover19a, kipf2016variational} encode the nodes of each graph into node-level embeddings and predict the links between each pair of nodes to generate a graph. 

In this work, we leverage recent advances in disentangled representation learning to further advance molecule generation. 

Currently, disentangled representation learning based on VAE is mainly limited in the domain of image representation learning~\cite{DBLP:conf/iclr/AlemiFD017, chen2018isolating,DBLP:conf/iclr/HigginsMPBGBML17,kim2018disentangling}. The goal is to learn representations that separate out the underlying explanatory factors responsible for formalizing the data. 
Disentangled representations are inherently more interpretable and can, thus, potentially facilitate debugging and auditing~\cite{DBLP:conf/iclr/AlemiFD017, chen2018isolating,esmaeili2019structured, kim2018disentangling,DBLP:conf/iclr/0001SB18,zhao2019infovae}. 

However, how to best learn representations that disentangle the latent factors behind a graph remains largely unexplored. Though few works are proposed for interpreting the graph representations~\cite{noutahi2019towards}, they do not focus on the graph generation task. In addition, utilizing disentanglement learning for molecule generation with desired properties is critical yet seldom explored. 
\section{Methods}
\label{sec:Methods}

\begin{figure*}[htbp]
  
  \includegraphics[width=\linewidth]{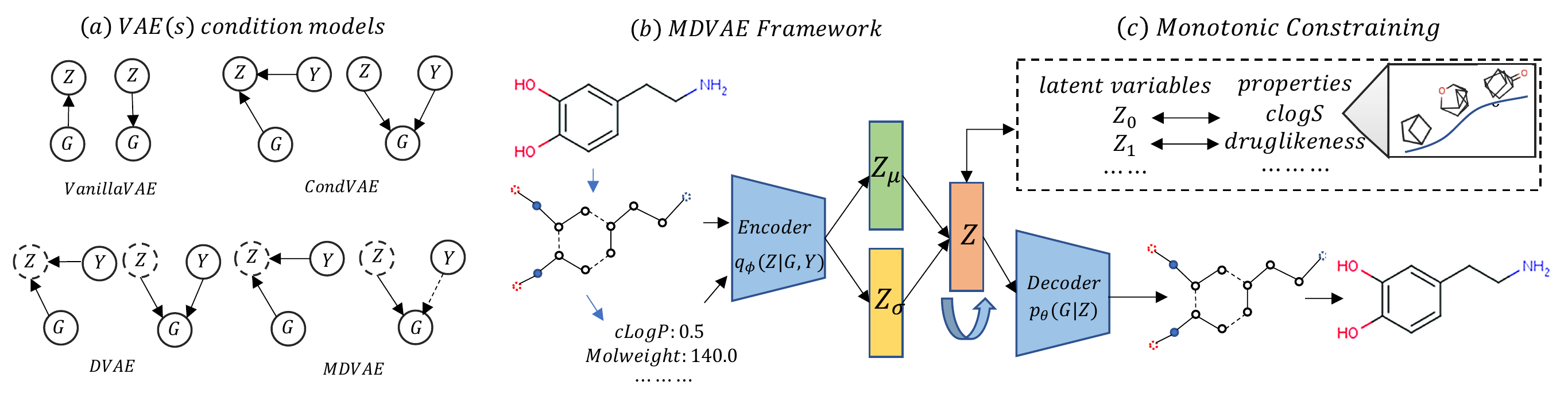}
  \caption{(a) The conditional models of the baseline models and our proposed models, including the encoder (left) and the decoder (right). The encoder encodes a molecular graph to a smooth latent representation. The decoder decodes a novel molecular graph from the latent representation (and a property Y). Dotted node represents disentangled latent representation, dotted line denotes polynomial and monotonic constraints. (b) MDVAE Framework. Molecule is represented as a graph, atom as a node, bond as a edge. The encoder encodes the graph into latent representation $z$, characterized by mean $z_{\mu}$ and $z_{\sigma}$, where disentalgment is enhanced. The decoder decodes the novel molecule from the latent representation. (c) Monotonic constraint is enforced in the latent representation $z$ to control the relationship between the latent variables and the observed properties. }
  \label{fig:diagram}
\end{figure*}

\subsection{Problem Formulation}
\label{subsec:ProblemFormulation}

The structure of a molecule can be defined as a graph  $G=(\mathcal{V},\mathcal{E},E, F)$, where $\mathcal{V}$ is the set of $N$ nodes (the atoms) and $\mathcal{E}\subseteq \mathcal{V} \times \mathcal{V}$ is the set of $M$ edges (the bonds that connect pairs of atoms). 
$e_{i,j}\in\mathcal{E}$ is an edge connecting nodes $v_i\in\mathcal{V}$ and $v_j\in\mathcal V$. 
$E\in \mathbb R^{ N\times N\times K}$ refers to the edge type tensor (i.e. bond type), where $E_{i,j}\in \mathbb R^{1\times K}$ is an one-hot vector encoding the type of edge $e_{i,j}$.
%$E_{i,j} \in \mathbb R^{1\times K }$ denotes the edge attributes of edge $e_{i,j}$ and where
$K$ is the number of edge types.
$F\in \mathbb R^{ N\times K'}$ refers to a node's feature matrix, where $F_{i}\in \mathbb R^{1\times K'}$ is the one-hot encoding vector denoting the type of atom $v_i\in\mathcal{V}$, and $K'$ is the total number of atom types. We also pre-define a set of molecular properties, such as clogP and molecular weight, as a vector of real-valued variables $Y=\{Y^{(1)},\ldots,Y^{(J)}\}$, where $Y^{(j)}$ is the value of the $j$-th property.
% based on deep generative models

Our goal is to learn the  generative process $p(G,Y|Z)$ of a molecule $G$ and its properties $Y$, characterized by latent variables $Z$. Considering that this process is obscure to be fully prescribed, we aim at characterizing it in a data-driven manner by latent variables $Z$ learned automatically by end-to-end deep generative models (e.g., VAEs and GANs). To further enhance the interpretability and controllability of $Z$, we will also aim to disentangle the latent variables and maximize the correspondence between (some of) them and the real molecule properties.

% poor understanding of the latent representation.  understand some of them are implicit  the others could be explicit and correspond to some predefined molecule properties. In the meanwhile,  better control different semantic factors (e.g. molecule properties) in formalizing a molecule. 
% The proposed problem goes beyond the existing molecule generation problem and requires enhanced interpretability and control over the generation process.

\subsection{Monotonically Disentangled VAE (MDVAE)}
\label{sec:OverallObjective}
Here we first introduce the proposed MDVAE model and its inference. Then we describe how to further enforce disentanglement among latent variables as well as their  monotonic relation to molecule properties.

\subsubsection{Disentangled Deep Generative Models for Molecule Graphs.}

Learning $p_{\theta}(G,Y|Z)$ requires the inference of its posterior $p_{\theta}(Z|G,Y)$. Because this inference is intractable, we need to define an approximated posterior $q_{\phi}(Z|G,Y)$ that is computationally tractable. We then minimize the Kullback–Leibler divergence (KLD) between them $D_{KL}(q_{\phi}(Z|G,Y)\|p_{\theta}(Z|G,Y))$ to ensure that the approximated posterior is close to the true posterior. This is well-known to amount to maximizing the evidence lower-bound~(\cite{bishop2006pattern}), as follows:
\begin{equation}
\begin{aligned}
\max_{\theta,\phi} \enspace & \mathbb{E}_{q_{\phi}(Z|G,Y)} [\log p_{\theta}(G,Y|Z)]\\
& -\lambda \sum\nolimits_i^N D_{\operatorname{KL}} (q_{\phi}(Z_i|G,Y) \| p(Z_i))
\end{aligned}
\end{equation}
\normalsize
where $\phi$ is the parameter of the approximated posterior $q_{\phi}$. The prior $p(Z)$ follows an isotropic Gaussian such that each $Z_i$ is independent of the others. Hence, the KL Divergence in the second term between $q_{\phi}(Z|G,Y)$ and $p(Z)$ will encourage the disentanglement among the variables in $Z$ in the inferred $q_{\phi}(Z|G,Y)$. Here $\lambda$ can control the strength of this enforcement, and the lager it is, the more independence different variables in $Z$ will have.
This can be achieved if we set the prior to be an isotropic unit Gaussian, i.e.~$p(Z)=\mathcal{N}(\mathbf{0},I)$, leading to the constrained optimization problem $\max_{\theta,\phi}\mathbb{E}_{G\thicksim \mathcal D}[\mathbb{E}_{q_{\phi}(Z|G,Y)}log p_{\theta}(G,Y|Z)]$ under the condition that $D_{KL}(q_{\phi}(Z|G,Y)||p(Z))\le\epsilon$,
where $\epsilon$ specifies the strength of the applied constraint; $\mathcal D$ refers to the observed set of molecules, and $D_{KL}(\cdot)$ denotes the Kullback–Leibler divergence (KLD) between two distributions. 
%As introduced in %DIP-VAE~\cite{kumar2017variational},
%the reconstruction error and KLD can be jointly minimized in the objective as:
%\small
%\begin{align}
%  &\max_{\theta} \mathbb{E}_{p_{\theta}(Z)} [\mathrm{log}p_{\theta}(G|Z)]-\lambda D_{KL}(q_{\phi}(Z|G)||p(Z))\\\nonumber
%  &s.t. \forall x_1\leq x_2:F_j(x_1)\leq F_j(x_2), \ \ x_1,x_2\sim q_\phi(Z^{(j)}|G),\ Z^{(j)}\subseteq Z\label{eq:2}
%\end{align}
%\normalsize
Considering $Z$ is conditionally independent to $Y$ given $G$ yields $q_{\phi}(Z|G,Y)=q_{\phi}(Z|G)$ and DIP-VAE~\cite{kumar2017variational} introduces disentanglement enforcement term $D(q_{\phi}(Z)||p(Z))$ with loss:
\begin{align} 
&\max_{\theta,\phi} \mathbb{E}_{q_{\phi}(Z|G)} [\mathrm{log}p_{\theta}(G|Z)]-\alpha D(q_{\phi}(Z)||p(Z))\label{eq:objective}\\\nonumber
+\mathbb{E}_{q_{\phi}(Z|G)}& [\mathrm{log}p_{\theta}(Y|G,Z)]-\lambda \sum\nolimits_i^N D_{KL}(q_{\phi}(Z_i|G)||p(Z_i))
\end{align}
which enforces the monotonicity of the relation between $Z$ and $Y$ in the following:
%Thus, we first decompose the second term in the objective as:
%\begin{align}\nonumber
%&-D_{KL}(q_{\phi}(Z|G)||p(Z))\\
%=& -\mathbb{E}_{q_{\phi}(Z,G)}log\frac{q_{\phi}(Z|G)}{q_{\phi}(Z)}-D_{KL}(q_{\phi}(Z)||p(Z)),
%\end{align}
%where the first term in the second row minimizes the mutual information $I(Z,G)$ in the inference model,
%while maximizing the second term (i.e., inferred priors) enforces the distance between $q_{\phi}(Z)$ and $p(Z)$.

%Since the first term actually represents the mutual information between the latent $Z$ and the graphs $G$, which will lead to poor reconstructions when enforcing disentanglement as mentioned in the first challenge. Thus, to solve the trade-off problems between the disentanglement of $Z$ and $G$, we discard it and maximize its lower bound instead as:
%\small
%\begin{align}\nonumber 
%&\ \ \ \ \ \max_{\theta} \mathbb{E}_{q_{\phi}(Z|G)} [\mathrm{log}p_{\theta}(G|Z)]-\lambda \sum_i^N D_{KL}(q_{\phi}(Z_i)||p(Z_i))\\\nonumber
%  &s.t. \forall x_1\leq x_2:F_j(x_1)\leq F_j(x_2), \ \ x_1,x_2\sim q_\phi(Z^{(j)}|G),\ Z^{(j)}\subseteq Z
%  \label{eq:objective}
%\end{align}\normalsize

%Considering that $Z=\{z_1,...,z_N\}$, then the \emph{Disentanglement Inferred Prior} term can be further written as:
%\begin{align}\nonumber
%  -D_{KL}(q_{\phi}(Z)||p(Z))&=
%  \mathbb{E}_{q_{\phi}(Z,G)}(log\frac{p(Z)}{q_{\phi}(Z)})\\\nonumber
%  = &\mathbb{E}_{q_{\phi}(Z,G)}(log\frac{\prod_i^N p(Z_i)}{\prod_i^N q_{\phi}(Z_i)}) \\
%  &=-\sum_i^N D_{KL}(q_{\phi}(Z_i)||p(Z_i)).
%\end{align}

\subsubsection{Monotonic Correlation towards Targeted Properties}

Disentanglement among latent variables $Z$ is deemed to improve the interpretability of VAE models~(\cite{Goodfellow-et-al-2016}). Here we go beyond this to correlate a subset of latent variables to important molecular properties (e.g., drug-likeness, water solubility, and clogP) to further enhance the interpretability of our latent variable, as well as better control the generated molecule's properties via tuning of the latent variables. 
% Here $Z'\in\mathbb{R}^{N\times H'}$ where $H'$ is the number of the latent factors in this subset of latent variables and we denote $Z'_i\in\mathbb\mathbb{R}^{1\times H'}$ as the latent variables for node $v_i$. 
% $T$ refers to the number of molecule properties.

Specifically, we have two aims: (1) \textbf{Aim 1: Correlating latent variables and real properties}: We explicitly relate one of the latent variables $Z_j\in Z$ in the disentangled latent representation to the predefined property set $Y_j$ in a pairwise style; (2) \textbf{Aim 2: Enforcing monotonic correlation}: We accommodate the non-linearity of the correlation between latent variables and targeted properties but encourage monotonic correlation, in order to ensure that the correlation is either positive or negative for effective control. This means that if we want to increase (or decrease) a given property, we can just increase (or decrease) the corresponding latent variable's value accordingly.

% For the first aim, we first consider the scenario where the molecule properties are independent with each other (the scenario where they are correlated will be discussed later). Here each property $Y_j$ corresponds to each latent variable $Z_j$, meaning $p(Y_j|Z)=p(Y_j|Z_j)$. This also leads to conditional independence between $G$ and $Y_j$ given $Z_j$, meaning $p_{\theta}(Y_j|G,Z_j)=p_{\theta}(Y_j|Z_j)$. The overall loss is expressed as:

For the first aim, we first consider the scenario where the molecule properties are independent with each other:

\begin{center}
\begin{align} 
&\max_{\theta,\phi} \mathbb{E}_{q_{\phi}(Z|G)} [\mathrm{log}p_{\theta}(G|Z)]-\alpha D(q_{\phi}(Z)||p(Z))\label{eq:objective2}\\\nonumber
&+\sum\nolimits_j^J\mathbb{E}_{q_{\phi}(Z|G)} [\mathrm{log}p_{\theta}(Y_j|Z_j)]\\\nonumber
&-\lambda \sum\nolimits_i^N D_{KL}(q_{\phi}(Z_i|G)||p(Z_i)) 
\end{align}
\end{center}
\normalsize 

where each property $Y_j$ corresponds to each latent variable $Z_j$, meaning $p(Y_j|Z)=p(Y_j|Z_j)$. This also leads to conditional independence between $G$ and $Y_j$ given $Z_j$, meaning $p_{\theta}(Y_j|G,Z_j)=p_{\theta}(Y_j|Z_j)$.

For the second aim, to enforce the monotonic relationship between properties and latent variables, we require for any property $j$ that we have
%Suppose there is a hidden function $F_j(\cdot)$ that instantiate this correlation: $y_j=F_j(Z^{(j)})$, then we enforce that this function is monotonic, namely 
$\forall Z_j^{(1)}\le Z_j^{(2)}: Y_j^{(1)}\le Y_j^{(2)} $, where $Z_j^{(1)},\ Z_j^{(2)}$ are two values of latent variable $Z_j$ from $q_\phi(Z_j|G)$, while $Y_j$ refers to any molecule property. The overall objective of our model is now reformulated as:

\begin{center}
\begin{align}
  \label{eq:3}
&\max_{\theta,\phi} \mathbb{E}_{q_{\phi}(Z|G)} [\mathrm{log}p_{\theta}(G|Z)] -\alpha D(q_{\phi}(Z)||p(Z))
\\\nonumber
&+ \sum\nolimits_j^J\mathbb{E}_{q_{\phi}(Z|G)} [\mathrm{log}p_{\theta}(Y_j|Z_j)\\\nonumber
&-\lambda \sum_i D_{KL}(q_{\phi}(Z_i|G)||p(Z_i)) \\\nonumber
  s.t.  \forall Z_j^{(1)}&\le Z_j^{(2)}: Y_j^{(1)}\le Y_j^{(2)},
  Z_j^{(1)},Z_j^{(2)}\sim q_\phi(Z_j|G).
\end{align}
\end{center}
\normalsize
As mentioned above, $Y_j$ is  dependent merely on $Z_j$, so we can define a function mapping $F_j:\mathbb{R}\rightarrow\mathbb{R}$ from $Z_j$ to $Y_j$. $F_j$ can be any function such as polynomial or multi-layer perceptron that can effectively fit arbitrarily complex (non-)linear mapping. Enforcing the constraint of Equation \eqref{eq:3} is equivalent to enforcing the monotonicity of function $F_j$. 

\subsection{Monotonic Regularization of MDVAE}
\label{sec:corr}
As mentioned in the discussion under Equation \eqref{eq:3}, we need to enforce the monotonicity of $F_j,\ j=\{1,\cdots,J\}$. This amounts to penalizing the violation of constraints via an additional regularization term $\mathcal{R}$ together with the original objective in Equation \eqref{eq:3}. In the following, we propose two different ways.

% Here, we propose two different ways to realize the monotonic regularization, namely, gradient-based and direction-based monotonic regularization. The gradient-based regularization explicitly enforces the monotonic relationship between the latent variables and properties by a monotonic non-linear polynomial function. The direction-based regularization does not require to learn an explicit function. It enforces the monotonic dependency between the properties and latent representation by directly enforcing the directions of variation of properties and latent variables to be the same. The details of these two implementations are as below. We adopt the gradient-based implementation in this paper to achieve better controllability by a stronger monotonic regularizer.

\noindent\textbf{Gradient-based monotonic regularization.}
In this way, we enforce that the gradient of function $F_j(Z_j)$ is always positive or negative to enforce its monotonicity. Without loss of generality, here we require the derivative to be always positive.
That means we want to enforce that $
    \frac{\mathrm{d}F_j(Z_j)}{\mathrm{d}Z_j}\ge 0
$, where $Z_j$ is any of the latent variables. Hence, the following regularization term will punish its violation: $\max (0,-\frac{\mathrm{d}F_j(Z_j)}{\mathrm{d}Z_j})$.

%That means we want to enforce $
%    (F_j(x_1)-F_j(x_2))\cdot(x_1-x_2)\ge 0
%$, where $x_1,\ x_2$ are any latent variables. Such nonlinear constraint is difficult to enforce so we instead penalize the following term in the objective as an equivalence:
%\begin{align}
 %   \max (0,-\frac{\mathrm{d}F_j(x)}{\mathrm{d}x})
%\end{align}
This term can be implemented using the following regularization term $\mathcal{R}(Z)$ using ReLU~\cite{you2018graph}, as in:
\begin{equation}
\label{eq:mono}
  \mathcal{R}(Z)=\sum\nolimits_j^J\mathbb{E}_{ q_\phi(Z|G)}\mathrm{ReLU}[-\frac{\mathrm{d}F_j(Z_j)}{\mathrm{d}Z_j}]
\end{equation}
\normalsize
where $J$ refers to the number of the targeted molecule properties. Note that without loss of generality, we only consider $Z_j$ as a scalar here. However, our framework can be easily extended to handle multiple latent variables by generalizing it as a vector; for each element in this vector, one can perform a ReLU operation the same as that in Eq.~\ref{eq:mono} and then sum up all of those corresponding to the multiple variables involved.

\noindent\textbf{Direction-based monotonic regularization.}
The second way starts from the standard definition of monotonicity: $
    (F_j(x_1)-F_j(x_2))\cdot(x_1-x_2)\ge 0
$, where $x_1,\ x_2$ are any latent variables. Such nonlinear constraint is difficult to enforce, so we instead penalize the following term in the objective as an equivalence: $\max (0,-(F_j(x_1)-F_j(x_2))\cdot(x_1-x_2))$, which can be implemented using the following regularization term $\mathcal{R}(y,Z)$ via ReLU function as well as additional denotations:

\begin{align}\nonumber
  \label{eq:6}
  \mathcal{R}(Y,Z)=&\sum\nolimits_j^J\mathbb{E}_{G_1,G_2\sim p_\theta(G|Z)}\\
  &[\sum\nolimits_i^N\mathrm{ReLU}[-(Y_j^{G_1}-Y_j^{G_2})
  (Z_j^{G_1}-Z_j^{G_2})],
\end{align}
%\normalsize
where $J$ refers to the number of the molecule properties, and $Y_j^{G_k}$ refers to the $j$-th molecule property of the molecule $G_k$. The molecule $G_1$ and $G_2$ are two arbitrary molecules sampled from the distribution of the observed graphs, and $Z_j^{G_1}$, $Z_j^{G_2}$ are $j$-th latent variable of them. Note that without loss of generality, our denotation here only considers $Z_j^{G_k}$ as a scalar, but our framework can be easily extended to handle multiple latent variables by generalizing it as a vector, similar to the Gradient-based approach.

% In this way, this term could punish the situation where the latent variable and the molecule factor of a pair of molecules do not change into the same direction (i.e. non-monotonous relationship), namely $(y_j^{G_1}-y_j^{G_2})(z_{i,j}^{G_1}-z_{i,j}^{G_2})<0$. The ReLu function is used to discover the pairs which meet the requirement. As a result, by adding this novel regularization term into the objective in Eq.~\ref{eq:objective}, the monotonic disentanglement molecule VAE (MD-MolVAE) is proposed. 

\begin{table*}[htbp]
  \centering
  \caption{Novelty, uniqueness, and validity are measured on molecule datasets generated by the various models under comparison. The highest value on a metric is highlighted in bold font.}
  \begin{adjustbox}{max width=\textwidth}
  \label{tab:freq}
  \begin{tabular}{c|ccccccccc}
    \hline\hline 
    Dataset & Metric & ChemVAE & GrammarVAE & GraphVAE & GraphGMG & LSTM & CGVAE & DVAE & MDVAE \\
    \hline
    \multirow{3}{*}{QM9} & \% Validity & 10.00 & 30.00 & 61.00 & - & 94.78 & \textbf{100.00} & \textbf{100.00} & \textbf{100.00}\\
    & \% Novelty & 90.00 & 95.44 & 85.00 & - & 82.98 & 96.33 &98.10 & \textbf{98.23}\\
    & \% Unique & 67.50 & 9.30 & 40.90 & - & 96.94 & 98.03 & 99.10 & \textbf{99.46} \\
    \hline
    \multirow{3}{*}{ZINC} & \% Validity & 17.00 & 31.00 & 14.00 & 89.20 & 96.80 & \textbf{100.00} & \textbf{100.00} & \textbf{100.00}\\
    & \% Novelty & 98.00 & \textbf{100.00} & \textbf{100.00} & 89.10 & \textbf{100.00} & \textbf{100.00} & \textbf{100.00} & \textbf{100.00} \\
    & \% Unique & 30.98 & 10.76 & 31.60 & 99.41 & 99.97 & 99.82 & 99.84 & \textbf{99.98}\\
    \hline\hline
\end{tabular}
\end{adjustbox}
\end{table*}

\noindent\textbf{Generalization of group-based correlated properties}.
To handle the situations when some molecule properties are correlated, we generalize the above framework to the group-based disentanglement strategy.

We define $j$-th group of variables $\mathcal{Z}_j\subseteq Z$ which correlate to a group of molecule properties $\mathcal{Y}_j\subseteq Y$.
Hence, the overall learning objective for the group-based disentanglement learning (Eq. \eqref{eq:3} + Eq. \eqref{eq:mono}) is:
\begin{align}
  \label{eq:final-objective}
&\max_{\theta, \phi} \mathbb{E}_{q_{\phi}(Z|G)} [\mathrm{log}p_{\theta}(G|Z)]-\alpha D(q_{\phi}(Z)||p(Z))\\\nonumber
&-\lambda\sum\nolimits_i^N D_{KL}(q_{\phi}(Z_i|G)||p(Z_i))\\\nonumber
&+\beta\sum\nolimits_j\sum\nolimits_i\mathbb{E}_{q_{\phi}(Z|G)} [\log p_\theta(\mathcal{Y}_{j,i}|\mathcal{Z}_{j})
\\\nonumber
&-\gamma\sum\nolimits_j\sum\nolimits_i\mathbb{E}_{ q_{\phi}(Z|G)}\mathrm{ReLU}[-\frac{\partial F_{j}(\mathcal{Z}_{j})}{\partial \mathcal{Z}_{j,i}}],  
\end{align}\normalsize

\section{Results}
\label{sec:Results}

This section reports on qualitative and quantitative experiments carried out to evaluate the performance of the proposed MDVAE model. All experiments are conducted on a 64-bit machine with an NVIDIA GPU (GeForce RTX 2080Ti, 1545MHz, 11GB GDDR6).

\paragraph{Experiment Set-up}
We compare MDVAE with $6$ state-of-the-art deep generative models on molecules: \emph{CGVAE}~\cite{NIPS2018_8005}, \emph{GraphGMG}~\cite{DBLP:journals/corr/abs-1803-03324}, \emph{SMILES-LSTM}~\cite{sundermeyer2012lstm}, \emph{ChemVAE}~\cite{gomez2018automatic}, \emph{GrammarVAE}~\cite{kusner2017grammar}, \emph{GraphVAE}~\cite{simonovsky2018graphvae}, detailed in Supplementary Material. We use the gradient-based approach for MDVAE during the experiments. Further, we add to this list the Disentangled VAE (DVAE) to serve as a baseline model. DVAE shares a similar objective with MDVAE but utilizes a linear reparametrization function rather than MDVAE's monotonic regularization term and polynomial reparametrization. It is worth noting that CGVAE~\cite{NIPS2018_8005} has a similar encoder and decoder to the proposed MDVAE and DVAE models but does not contain the proposed disentanglement and monotonic enforcement. So, the comparison of MDVAE with CGVAE represents an ablation study that allows us to test the effectiveness of the proposed disentanglement and monotonic enforcement in MDVAE and the comparison of MDVAE with DVAE represent an ablation study that test the effectiveness of polynomial reparametrization and monotonic enforcement. Detailed model hyperparameters and architectures can be found in Supplementary Material.

\paragraph{Datasets}

We consider two popular benchmark datasets. (1) The \emph{QM9 Dataset}~\cite{ramakrishnan2014quantum} consists of around 134k stable small organic molecules with up to $9$ heavy atoms (Carbon (C), Oxygen (O), Nitrogen (N) and Fluorine (F)), with a 120k/20k split for training versus validation. (2) The \emph{ZINC Dataset}~\cite{irwin2012zinc} contains around 250k drug-like chemical compounds with an average of around $23$ heavy atoms, with a 60k/10k split for training versus validation.

\subsection{Comparison with State-of-the-art Methods}

We first evaluate and compare the quality of generated molecules across the various deep generative models. All models are trained on each of the two benchmark datasets, and $30,000$ molecules are then generated/sampled from each trained model for the purpose of evaluation. 

Table~\ref{tab:freq} reports on three popular metrics: \emph{novelty}, which  measures the fraction of generated molecules that are not in the training dataset; 
\emph{uniqueness}, which measures the fraction of generated molecules after and before removing duplicates; and \emph{validity}, which measures the fraction of generated molecules that are chemically valid. As Table~\ref{tab:freq} shows, CGVAE, MDVAE, and DVAE achieve $100\%$ validity; that is, $100$\% of generated molecules are chemically-valid, which is significantly higher than other methods. This is due to the sequence decoding process, which takes a valency check step by step and so ensures that generated molecules are valid. 

Table~\ref{tab:freq} also shows that MDVAE and DVAE generate up to $100\%$ novel molecules, which is higher than other methods, including CGVAE. Note that CGVAE shares a similar architecture with MDVAE and DVAE but without the disentanglement enforcement. This allows us to conclude that the higher novelty achieved by MDVAE and DVAE is due to the disentangled representation, which can fully explore molecular patterns. In particular, adding the disentanglement regularization does not affect the reconstruction error and so does not sacrifice the quality of generated molecules. We note that the LSTM method also works well on the ZINC dataset but worse on the QM9 dataset, which has a more complex data distribution with drug-like molecules. Our models and CGVAE have the highest performance on uniqueness, over $99$\%. 
% Randomly selected molecules generated by CGVAE, DVAE, and MDVAE are shown in the Supplemental Material. 

\begin{table}[htbp]
\small
  \centering
  \caption{Evaluation of disentanglement on the QM9 and ZINC datasets.}
  \label{tab:disentangle_prop}
  \begin{adjustbox}{max width=0.48\textwidth}
  \begin{tabular}{|c|ccccc|}
    \hline
    Dataset&Model&$\beta$-M (\%)$\uparrow$&F-M(\%)$\uparrow$&DCI$\uparrow$&Mod$\uparrow$\\
    \hline
    \multirow{3}{*}{QM9} & CGVAE & \textbf{100} & 72.0 & 0.151 & 0.634\\
    & DVAE & \textbf{100} & 76.4 & 0.152 & 0.671 \\
    & MDVAE & \textbf{100} & \textbf{78.8} & \textbf{0.209} & \textbf{0.690}\\
    \hline
    \multirow{3}{*}{ZINC} & CGVAE & \textbf{100} & 61.6 & 0.109 & 0.604\\
    & DVAE & \textbf{100} & 62.4 & 0.111 & 0.611\\
    & MDVAE & \textbf{100} & \textbf{64.4} & \textbf{0.156} & \textbf{0.621}\\
    \hline
\end{tabular}
\end{adjustbox}
\end{table}

\begin{figure*}[htbp]
  \centering
% \begin{tabular}{c}
  \includegraphics[width=0.8\linewidth]{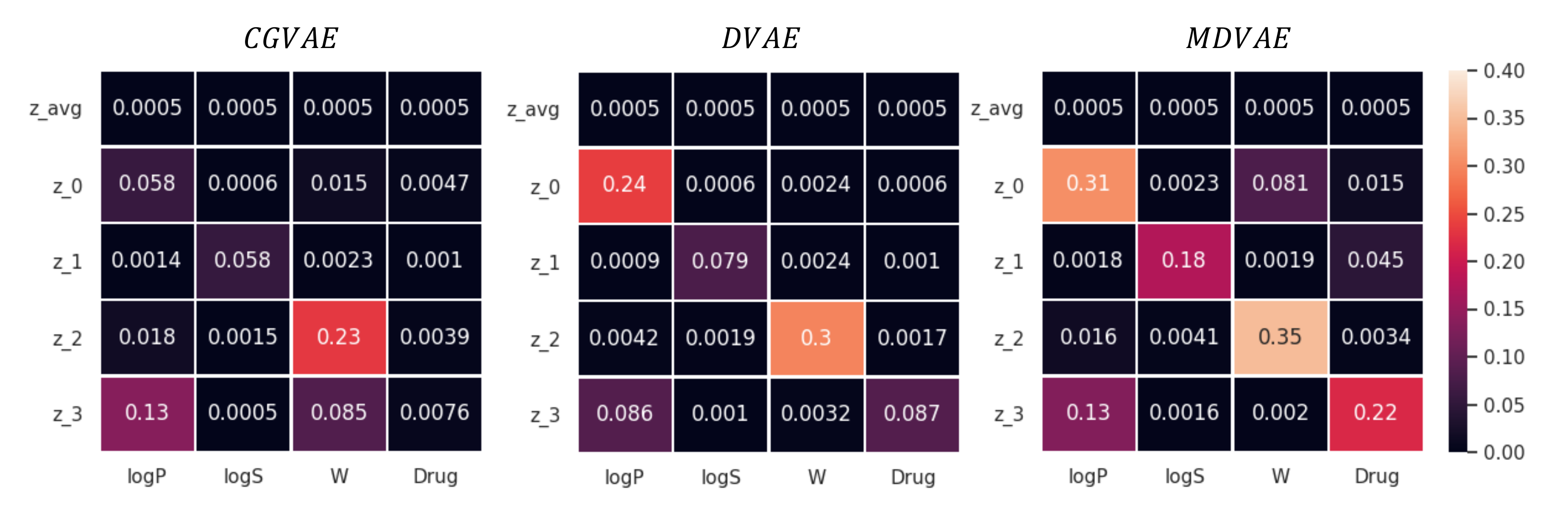}\\
%   \end{tabular}
  \caption{MI heatmaps between latent variables and molecular properties (cLogP, cLogS, molecular weight, and drug-likeness. MI is normalized between 0 and 1) (better seen in color).}
  \label{fig:prop_control}
\end{figure*}

\subsection{Evaluating Impact of Disentanglement}

We now further compare CGVAE, DVAE, and MDVAE in Table \ref{tab:disentangle_prop} on disentanglement of the learned latent distributions. We utilize four popular metrics to do so: \textit{$\beta$-M}\cite{DBLP:conf/iclr/HigginsMPBGBML17}, \textit{F-M}\cite{kim2018disentangling}, \textit{MOD}~\cite{ridgeway2018learning}, \textit{DCI}~\cite{eastwood2018framework}, detailed explanations can be found in Supplementary Material. 

% \textit{$\beta$-M}\cite{DBLP:conf/iclr/HigginsMPBGBML17}
% measures disentanglement by examining the accuracy of a linear classifier that predicts the index of a fixed factor of variation.
% \textit{F-M}\cite{kim2018disentangling} addresses several issues by using a majority voting classifier
% on a different feature vector that represents a corner case in the $\beta$-M.
% \textit{MOD}~\cite{ridgeway2018learning} measures whether each latent variable depends on at most a factor describing the maximum variation using their mutual information.
% \textit{DCI}~\cite{eastwood2018framework} computes the entropy of the distribution obtained by normalizing the importance of each dimension of the learned representation for predicting the value of a factor of variation. All implementation details are in the work proposed by~\cite{locatello2018challenging}.

Table \ref{tab:disentangle_prop} shows that our models achieve the best overall disentanglement scores. Specifically, all models achieve $100$ on the $\beta$-M score on both datasets. On the F-M score, on both datasets, MDVAE performs best, followed by DVAE, and CGVAE in this order. Similar ranking is observed on the DCI and Mod scores on both datasets. On the DCI score, DVAE performs slightly higher than CGVAE, with MDVAE performing much better than both. Similar observations hold on the Mod score comparison. It is worth noting that all three models are challenged more by the ZINC than the QM9 dataset with regards to the DCI score; the ZINC dataset is larger and contains larger molecules, as well, and it is possible that this results in more latent variables. Altogether, these results show that the proposed MDVAE and DVAE can successfully learn the disentangled latent representations better than CGVAE.

\subsection{Performance in Molecular Property Control}

Here, CGVAE, DVAE, and MDVAE are evaluated on whether the latent variables capture desired properties. 
% First, we show the relationship between pairs of nine molecular properties by evaluating the linear correlation of pairs in the heatmap in Supplementary Material Fig~\ref{fig:heatmap}. 
First, we identify four properties, cLogP, cLogS, molecular weight, and drug-likeness to evaluate further (the linear correlation of paired molecular properties is visualized in Supplementary Material; specifically, we utilize mutual information (MI), implemented via the the scikit-learn library, to measure the mutual dependency between each latent variable and each of the four above properties. These results are related via heatmaps in Figure~\ref{fig:prop_control}, which shows the MI for the pairs ($z_{0}$, clogP), ($z_{1}$, cLogS), ($z_{2}$, molecular weight), and ($z_{3}$, drug-likeness). To make CGVAE comparable to our polynomial reparameterization and monotonic constraint, we implement a linear control over the latent representation $z$ and properties $p$. It is clear that the polynomial function greatly increases property control; a much larger MI is achieved with all four properties. DVAE ranks second, which shows that disentanglement enhancement improves controllability. The other latent variables do not interfere with the properties with which we pair the specific latent variables $z_{0}-z_{4}$. The conditioning on drug-likeness has a strong control over the cLogP property; even though we do not observe a linear correlation between these two properties, drug discovery literature shows that drug-likeness is correlated to cLogP~\cite{leeson2007influence}.

\subsection{Qualitative Evaluation for Disentanglement and Property Control}

\begin{figure}[htbp]
\centering
\begin{tabular}{cc}
\includegraphics[width=0.43\columnwidth]{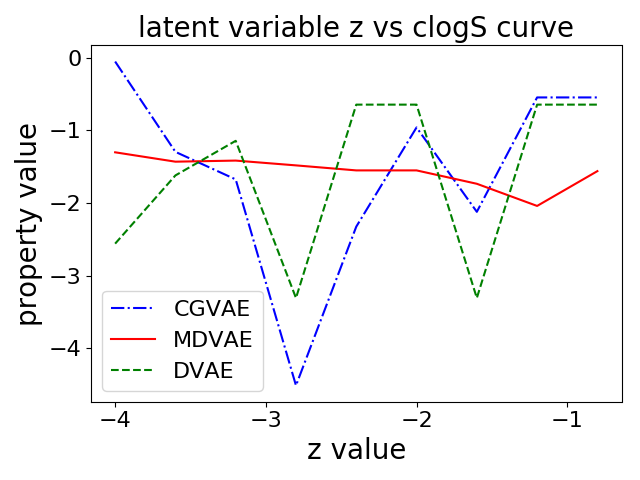} &
\includegraphics[width=0.43\columnwidth]{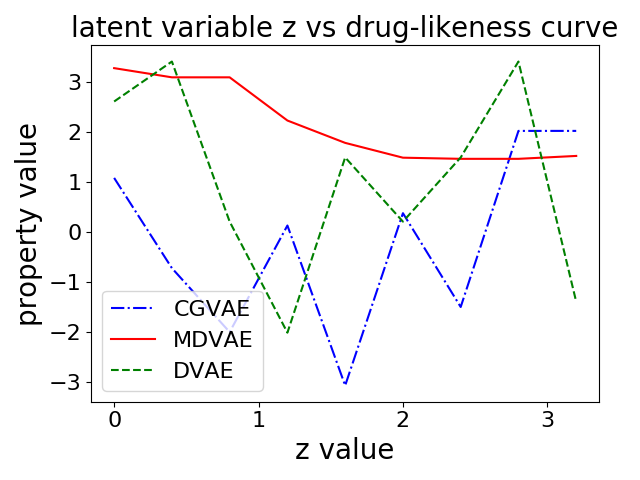}\\
\includegraphics[width=0.43\columnwidth]{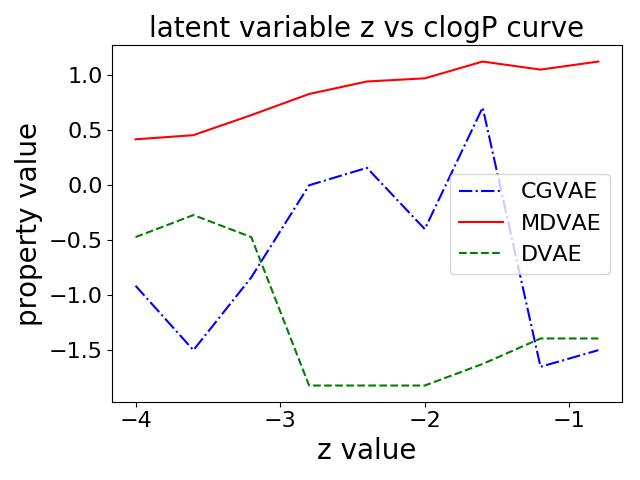} &
\includegraphics[width=0.43\columnwidth]{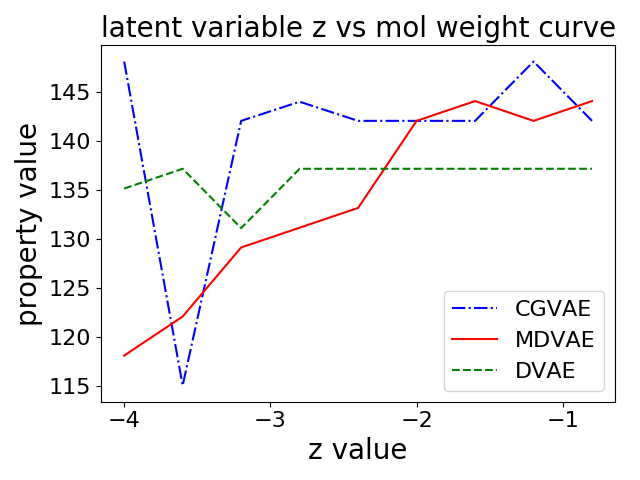}\\
\end{tabular}
\caption{The relationship between the latent variable $z_i$ and properties of the molecules generated by CGVAE, DVAE, and MDVAE.}
\label{fig:relation_z_factor}
\end{figure}

We demonstrate qualitatively that MDVAE and DVAE consistently discover latent variables and use them to control molecular properties in a monotonous fashion. By jointly changing the value of one latent variable continuously and fixing the remaining latent variables, we can visualize the corresponding variation of molecular properties in the generated graphs. Figure~\ref{fig:relation_z_factor} plots the variation of each property along with the change of its target latent variable. More results can be found in Supplementary Material.

Figure~\ref{fig:relation_z_factor} shows that MDVAE monotonically captures the disentangled latent variables that control molecular weight and drug-likeness. There are obvious fluctuations of all four properties when controlled by the latent variables learned from DVAE and CGVAE (see the green and blue lines). This is most visible in the steep decrease and increase of clogS when $z$ ranges from $-4$ to $-1$, while the property of the generated molecules by MDVAE monotonically decreases (see the red solid line). This demonstrates that the proposed monotonic correlation regularization term is necessary and effective in preserving the monotonic correlation between each molecule property and its relevant latent variable for better control of the molecule generation to obtain the desired properties. 

\begin{figure}[htbp]
\centering
\begin{tabular}{c}
\includegraphics[height=0.38\columnwidth]{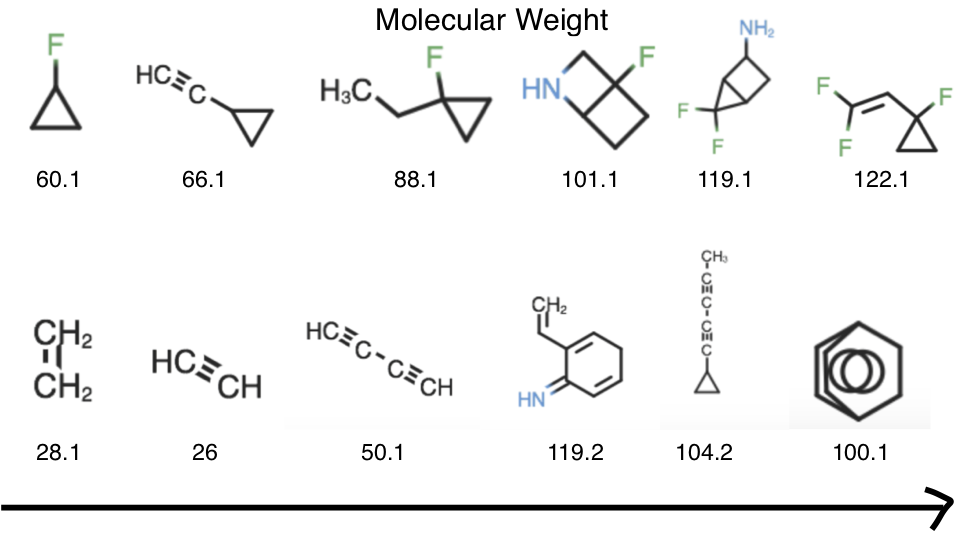}
%\\
%\includegraphics[height=0.40\columnwidth]{Figures/Mol_viz_drug.png}
\end{tabular}
\caption{Molecules generated by MDVAE (top row) and DVAE (bottom row) as we increase the value of the latent variable for molecular weight.}
\label{fig:disentangle_MV_viz}
\end{figure}

Finally, in Figure~\ref{fig:disentangle_MV_viz} we show how generated molecules change when the value of the latent variable $z_{0}$ and $z_{1}$ changes from $-5$ to $5$ and $5$ to $-5$. The molecular weight scores are shown at the bottom of each molecule. Compared to DVAE, the proposed MDVAE model is more powerful at generating valid and high-quality molecules along with the variation of the latent variables.  We can observe that molecular weight increases with the increase of the value of one of the latent variables. More results can be found in Supplementary Material.

\section{Conclusion}
\label{sec:conclusion}

This paper proposes a new disentangled deep generative framework for interpretable molecule generation with property control via a graph-based disentangled VAE. We derive new objectives which further enforce non-linearity and monotonicity of the relation between some latent variables and target molecule properties. The proposed models are validated on two real-world molecule datasets for three tasks: molecule generation, disentangled representation learning, and control of the generation process. Quantitative and qualitative evaluation results show the promise of disentangled representation learning in interpreting and controlling molecular properties during the generation process. 

\bibliographystyle{unsrt}
\bibliography{sdm22}

\clearpage
\appendix

\section{Supplemental Material}

\subsection{Comparison Methods}
\label{sec:comparison}
MDVAE is compared to $6$ state-of-the-art deep generative frameworks: 
(1) \emph{CGVAE}~\cite{NIPS2018_8005}, a VAE-based model consisting of a graph-structured encoder and decoder;
(2) \emph{GraphGMG}~\cite{DBLP:journals/corr/abs-1803-03324}, a deep auto-regressive graph model that generates graph nodes sequentially;
(3) \emph{SMILES-LSTM}~\cite{sundermeyer2012lstm}, an LSTM model that utilizes the SMILES representation;
(4) \emph{ChemVAE}~\cite{gomez2018automatic}, a generative model that converts discrete representations of molecules to and from a multidimensional continuous representation;
(5) \emph{GrammarVAE}~\cite{kusner2017grammar}, a VAE-based model that enforces syntactic and semantic constraints over SMILES strings via context free and attribute grammars;
and (6) \emph{GraphVAE}~\cite{simonovsky2018graphvae}, a generic deep generative model for graph generation.

% \subsection{Pairwise Correlation of Molecular Properties}

% We show the pairwise correlation between some selected common molecular properties in Fig. \ref{fig:heatmap}. We can see that, molecular weight and drug-likeness poses the least linear correlation with each other. We test both the correlated properties and the uncorrelated properties in our experiments.

% \begin{figure}[htp]
%     \centering
%     \includegraphics[width=0.45\textwidth]{Figures/heatmap.pdf}
%     \caption{Heatmap showing the linear correlation between each pair of the given properties. Each entry is the correlation score between a pair of properties of molecule.}
%     \label{fig:heatmap}
% \end{figure}

% \subsection{Visualization of Randomly Selected Molecules}

% Figure~\ref{fig:mol_viz} shows some molecules from the QM9 dataset (top panel) and some molecules generated by CGVAE, DVAE, and MDVAE (the other panels). All the molecules are selected randomly. 

% \begin{figure}[htbp]
%     \centering
%     \includegraphics[width=0.43\textwidth]{Figures/Mol_viz_pic.png}
%     \caption{Some molecules from QM9 dataset are shown on the top panel. The other panels show some molecules generated by CGVAE, DVAE, and MDVAE, respectively.}
%     \label{fig:mol_viz}
% \end{figure}

\subsection{Metrics to Measure Disentanglement} 
\label{suppmetric}

\textit{$\beta$-M}\cite{DBLP:conf/iclr/HigginsMPBGBML17}
measures disentanglement by examining the accuracy of a linear classifier that predicts the index of a fixed factor of variation.
\textit{F-M}\cite{kim2018disentangling} addresses several issues by using a majority voting classifier
on a different feature vector that represents a corner case in the $\beta$-M.
\textit{MOD}~\cite{ridgeway2018learning} measures whether each latent variable depends on at most a factor describing the maximum variation using their mutual information.
\textit{DCI}~\cite{eastwood2018framework} computes the entropy of the distribution obtained by normalizing the importance of each dimension of the learned representation for predicting the value of a factor of variation. All implementation details are in the work proposed by~\cite{locatello2018challenging}.

% \subsection{Architecture and Hyper-parameters}
% The deatiled implementation of the architecture is based on the work by~\cite{liu2018constrained}. The structure of a molecule can be defined as a graph  $G=(\mathcal{V},\mathcal{E},E, F)$, where $\mathcal{V}$ is the set of $N$ nodes (the atoms) and $\mathcal{E}\subseteq \mathcal{V} \times \mathcal{V}$ is the set of $M$ edges (the bonds that connect pairs of atoms). 
% $e_{i,j}\in\mathcal{E}$ is an edge connecting nodes $v_i\in\mathcal{V}$ and $v_j\in\mathcal V$. 
% $E\in \mathbb R^{ N\times N\times K}$ refers to the edge type tensor (i.e. bond type), where $E_{i,j}\in \mathbb R^{1\times K}$ is an one-hot vector encoding the type of edge $e_{i,j}$.
% %$E_{i,j} \in \mathbb R^{1\times K }$ denotes the edge attributes of edge $e_{i,j}$ and where
% $K$ is the number of edge types.
% $F\in \mathbb R^{ N\times K'}$ refers to a node's feature matrix, where $F_{i}\in \mathbb R^{1\times K'}$ is the one-hot encoding vector denoting the type of atom $v_i\in\mathcal{V}$, and $K'$ is the total number of atom types. We also pre-define a set of molecular properties, such as clogP and molecular weight, as a vector of real-valued variables $Y=\{Y^{(1)},\ldots,Y^{(J)}\}$, where $Y^{(j)}$ is the value of the $j$-th property. 
\begin{table}[htb]\small
\caption{Encoders and decoders architectures of MDVAE for QM9 and ZINC dataset (Each layers is expressed in the format as $<kernel\_size><layer\_type><Num\_channel><Activation\_function><stride\_size>$. \textit{FC} refers to the fully connected layers).}
    \centering
    \begin{adjustbox}{max width=0.48\textwidth}
    \begin{tabular}{|l|l|l|}
    \hline
        Encoder&Decoder\\\hline
        Input: $G (\mathcal{V},\mathcal{E},E, F)$&Input$[z]\in \mathbb{R}^{100}$  \\\hline
       FC.100 ReLU& FC.100 ReLU\\\hline 
       GGNN.100 ReLU & GGNN.100 ReLU\\\hline
       GGNN.100 ReLU & GGNN.100 ReLU\\\hline
       FC.100 & FC.bv (batch node size) FC.3 (edge) \\\hline
    \end{tabular}
    \label{tab:details of MDVAE}
    \end{adjustbox}
\end{table}

\begin{table}[htb]\small
\caption{hyper-paramter used for training on dSprites and QM9 datasets}
    \centering
    \begin{adjustbox}{max width=0.48\textwidth}
    \begin{tabular}{|l|l|l|l|l|l|l|}
    \hline
     Dataset&Learning\_rate&Batch\_size&$\lambda$&Num\_iteration\\\hline
       QM9&5e-4&64&1&10\\\hline
       ZINC&5e-4&8&1&5\\\hline
    \end{tabular}
    \label{tab:details of training}
    \end{adjustbox}
\end{table}

\subsection{Model Complexity Analysis}
The proposed DVAE and MDVAE models share the same time complexity with the baseline CGVAE model, which is based on a typical Gated Graph Neural Network (GNN) to encoder the molecules and is of linear growth with respect to the number of node $N$, and the decoder part is formulated as a sequential generation process, where the worst case time complexity is $O(N^2)$. Overall, our models amount to a worst case $O(N^2)$ time complexity, which is scalable compared to the state-of-the-art graph generative models, worst case $O(N^4)$ for graphVAE \cite{simonovsky2018graphvae} and worst case $O(N^2)$ for graphRNN \cite{you2018graphrnn}.

\begin{figure}[htp]
    \centering
\begin{tabular}{c}
    \includegraphics[width=0.45\textwidth]{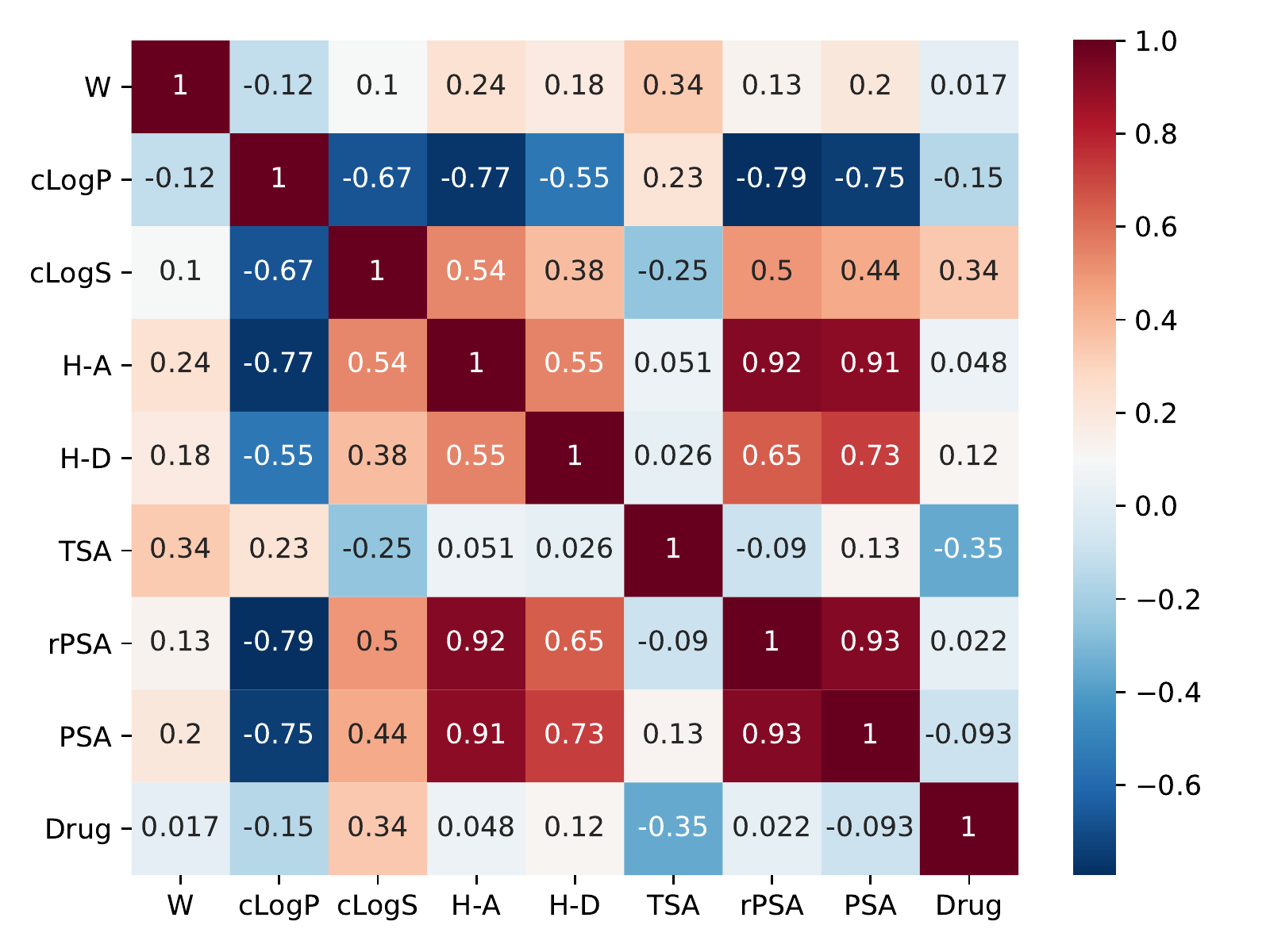}\\
    \end{tabular}
    \caption{Heatmap showing the linear correlation between each pair of the given properties. Each entry is the correlation score between a pair of properties of molecule.}
\end{figure}

\subsection{Implementation of MDVAE} The construction of the overall model extends the conventional VAE consisting of an encoder and a decoder, where the former implements the inference model $q_{\phi}(Z|G)$ while the latter instantiates the generator $p_{\theta}(G|Z,Y)$. 

\noindent\textbf{Molecule Encoder} To model the prior distributions $q_\phi(Z|G)$ expressed in the objective, an encoder is constructed based on a graph neural network (GNN).  The GNN embeds each node in an input graph $G$ into $L$-dimension latent space following distribution $q_\phi(Z|G)$ parameterized by mean $\mu_i$ and standard deviation vectors $\sigma_i$ for each node $v_i$, which is the output of the GNN. Note the model handles the inputs of different sizes by setting a maximum number of vertices and mask on the padded nodes in each batch. To compensate the changing maximum vertices for each batch, the model is designed to pass into the network by the hidden size (h) rather than number of vertices  $\times$ the hidden size ($v \times h$) as other regular GNN does. In the end, by sampling from the modeled distribution, $Z=\{Z_1,...,Z_N\}$ are obtained variables containing the node representation vectors for all the nodes.

\noindent\textbf{Molecule Decoder} The molecule decoder models the distribution $p_{\theta}(G|Z)$ by generating the molecule graph $G$ conditioning on the latent representation variables $Z$ that are sampled from the learned distribution in the encoder. The process proceeds in an auto-regressive style. In each step a focus node is chosen to be visited, and then the edges are generated related to this focus node are generated. The nodes are ordered by using the breadth-first traversal. The molecule decoder mainly contains three steps, namely \emph{node initialization}, \emph{node update} and \emph{edge selection and labeling}. The detailed implementation of the encoder and decoder is based on the work by~\cite{liu2018constrained}.

% \subsection{Model Complexity Analysis}
% The proposed DVAE and MDVAE models share the same time complexity with the baseline CGVAE model, which is based on a typical Gated Graph Neural Network (GNN) to encoder the molecules and the decoder part is formulated as a sequential generation process. In the worst case, time complexity is $O(N^2)$, which is scalable compared to the state-of-the-art graph generative models, worst case $O(N^4)$ for graphVAE \cite{simonovsky2018graphvae} and worst case $O(N^2)$ for graphRNN \cite{you2018graphrnn}.

\subsection{Hyperparameter Details}
\begin{table}[htb]\small
\caption{Encoders and decoders architectures of MDVAE for QM9 and ZINC dataset (Each layers is expressed in the format as $<kernel\_size><layer\_type><Num\_channel><Activation\_function><stride\_size>$. \textit{FC} refers to the fully connected layers).}
    \centering
    \begin{adjustbox}{max width=0.48\textwidth}
    \begin{tabular}{|l|l|l|}
    \hline
        Encoder&Decoder\\\hline
        Input: $G (\mathcal{V},\mathcal{E},E, F)$&Input$[z]\in \mathbb{R}^{100}$  \\\hline
       FC.100 ReLU& FC.100 ReLU\\\hline 
       GGNN.100 ReLU & GGNN.100 ReLU\\\hline
       GGNN.100 ReLU & GGNN.100 ReLU\\\hline
       FC.100 & FC.bv (batch node size) FC.3 (edge) \\\hline
    \end{tabular}
    \label{tab:details of MDVAE}
    \end{adjustbox}
\end{table}

\begin{table}[htb]\small
\caption{hyper-paramter used for training on dSprites and QM9 datasets}
    \centering
    \begin{adjustbox}{max width=0.48\textwidth}
    \begin{tabular}{|l|l|l|l|l|l|l|}
    \hline
     Dataset&Learning\_rate&Batch\_size&$\lambda$&Num\_iteration\\\hline
       QM9&5e-4&64&1&10\\\hline
       ZINC&5e-4&8&1&5\\\hline
    \end{tabular}
    \label{tab:details of training}
    \end{adjustbox}
\end{table}

\subsection{Comparison of the Learned to the Reference Distribution}

The above analysis indicates that CGVAE, DVAE, and MDVAE are the top three models, so we now focus on further evaluation of these three models. In particular, we now compare the reference/training dataset to the dataset generated by a model. Each dataset is summarized via a distribution of a variable of interest. The divergence between two distributions is measured via popular metrics, such as the Maximum Mean Discrepancy (MMD)~\cite{you2018graphrnn} and the Kullback-Leibler Divergence (KLD)~\cite{you2018graphrnn}. 

\begin{table}[htb]\small
  \centering
  \caption{Comparison of training and generated distributions of graph properties via MMD and KLD. (CC refers to the Clustering Coefficient).}
  \label{tab:graph_prop}
  \begin{adjustbox}{max width=0.48\textwidth}
  \begin{tabular}{|c|c|ccc|}
    \hline
    Dataset&Metric& CGVAE & DVAE & MDVAE \\\hline
     \multirow{8}{*}{QM9}&MMD(Degree)& 0.0985 & 0.1133 & \textbf{0.0456}\\
     &MMD (CC)& 0.0215 & 0.0242 & \textbf{0.0192} \\
     &MMD (Orbit)& 0.0179 & 0.0130 & \textbf{0.0057}\\
     &KLD (cLogP) & 0.3102 & 0.2744 & \textbf{0.1935}\\
     &KLD (cLogS) & 0.2886 & 0.2297 & \textbf{0.1358}\\
     &KLD (Drug-like) & 0.0100 & 0.1261 & \textbf{0.0099}\\
     &KLD (PSA) & 0.3921 & 0.2931 & \textbf{0.2277}\\
     &KLD (Mol Weight) & 3.1360 & 1.7641 & \textbf{1.7005}\\
    \hline
     \multirow{8}{*}{ZINC}&
      MMD (Degree)& 0.0050 & 0.0057 & \textbf{0.0045}\\
      &MMD (CC))& 0.0012 & 0.0008 & \textbf{0.0004} \\
     &MMD (Orbit)& 0.0011 & 0.0018 & \textbf{0.0011}\\
     &KLD (cLogP) & 0.1888 & 0.0503 & \textbf{0.0428}\\
     &KLD (cLogS) & 0.0641 & 0.0494 & \textbf{0.0483}\\
     &KLD (Drug-like) & 1.4590 & 0.7671 & \textbf{0.3268}\\
     &KLD (PSA) & 0.1918 & 0.3159 & \textbf{0.1686}\\
     &KLD (Mol Weight) & 0.1191 & 0.1296 & \textbf{0.1115}\\
    \hline
\end{tabular}
\end{adjustbox}
\end{table}

We utilize three popular graph-summarization variables~\cite{you2018graphrnn,liu2018constrained}: \emph{node degree}, \emph{clustering coefficient}, and \emph{average orbit count}; the latter counts the number of $4$-orbits in a graph. Distributions are compared via MMD. In addition, we utilize benchmark molecular properties utilized in medicinal chemistry to evaluate molecular compounds, such as cLogP, cLogS, Drug-likeness, Polar Surface Area (PSA), and Molecular Weight~\cite{NIPS2018_8005} and use KLD to compare resulting distributions. Briefly, cLogP measures hydrophilicity and evaluates the absorption or permeation of a compound; cLogS provides complementary information on aquous solubility; Drug-likeness is partially based on topological descriptors; PSA evaluates the cell permeability of a compound; and molecular weight provides information on both compound activity and permeability.

Table~\ref{tab:graph_prop} shows that on the QM9 dataset, the MDVAE-generated dataset is more similar to the reference dataset than the DVAE- and CGVAE-generated datasets; the smaller the value is, the more similar the generated properties are to those in the training set. Similarly, all MMD values are reasonably small, suggesting that DVAE and MDVAE also preserve graph patterns of the reference distribution. All three models perform poorly on the molecular weight property, especially on the QM9 drug-like molecule dataset. We suspect the reason is that the drug-like molecule distribution is harder to learn. Comparably, all three models preserve the reference distribution more on the ZINC than on the QM9 dataset. 

Figure~\ref{fig:mol_prop} provides some more detail and relates the distribution of generated molecules superimposed over the distribution of the reference dataset (in terms of selected molecular properties). Figure~\ref{fig:mol_prop} shows that there is good agreement between the generated and reference distributions. Of all the three models, MDVAE performs best, yielding greatest overlap between the generated data property distribution and the reference dataset property distribution. Nevertheless, DVAE and CGVAE perform comparably well, with DVAE slightly outperforming CGVAE. 

% \subsection{Performance in Molecular Property Control}
% \label{sec:supp_control}
% \begin{figure}[htbp]
% \centering
% \begin{tabular}{cc}
% \includegraphics[width=0.45\columnwidth]{Figures/clogS_control.png} &
% \includegraphics[width=0.45\columnwidth]{Figures/druglikeness_control.png}
% \end{tabular}
% \caption{The relationship between the latent variable $z_i$ and properties of the molecules generated by CGVAE, DVAE, and MDVAE.}
% \end{figure}

\begin{figure}[htbp]
\centering
\begin{tabular}{c}
% \includegraphics[height=0.40\columnwidth]{Figures/Disentangled_Mol_Viz.png}
% \\
\includegraphics[height=0.48\columnwidth]{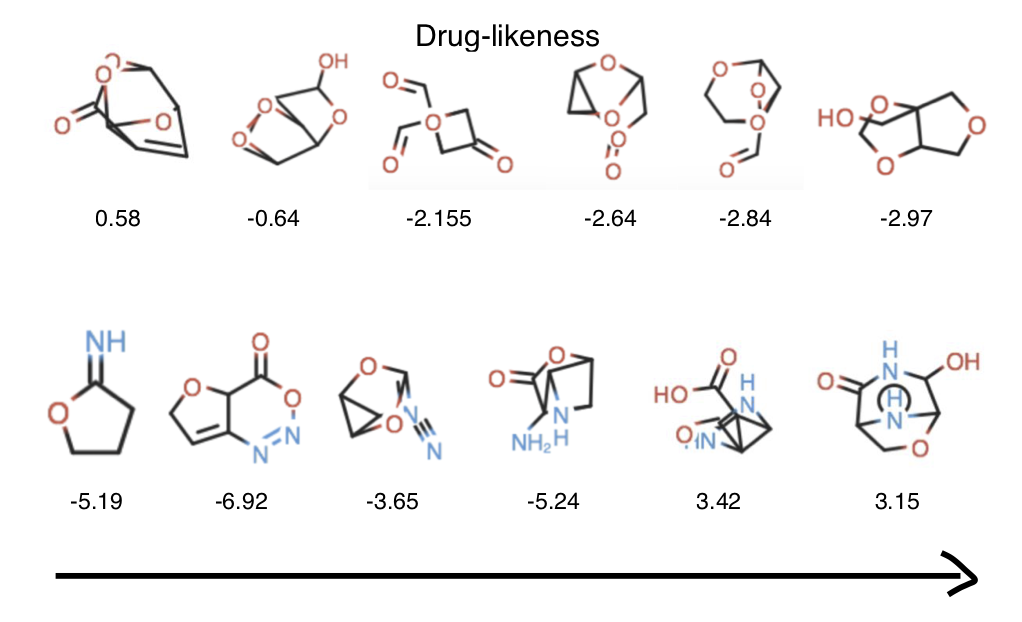}
\end{tabular}
\caption{Molecules generated by MDVAE (top row) and DVAE (bottom row) as we increase the value of the latent variable by enforcing the molecular weight (left) and drug-likeness (right).}
\end{figure}

% \begin{figure}[htbp]
% \centering
% \begin{tabular}{cc}    
% \multicolumn{2}{c}{\textbf{CGVAE}}\\
% clogP&clogS\\
% \includegraphics[width=0.23\textwidth]{Figures/CGVAE_cLogP.png} &
% \includegraphics[width=0.23\textwidth]{Figures/CGVAE_cLogS.png} \\
% Mol weight&Drug-likeness\\
% \includegraphics[width=0.23\textwidth]{Figures/CGVAE_Molweight.png} &
% \includegraphics[width=0.23\textwidth]{Figures/CGVAE_Druglikeness.png} \\
% \multicolumn{2}{c}{\textbf{DVAE}}\\
% \includegraphics[width=0.23\textwidth]{Figures/DVAE_cLogP.png} &
% \includegraphics[width=0.23\textwidth]{Figures/DVAE_cLogS.png} \\
% Mol weight&Drug-likeness\\
% \includegraphics[width=0.23\textwidth]{Figures/DVAE_Molweight.png} &
% \includegraphics[width=0.23\textwidth]{Figures/DVAE_Druglikeness.png}\\
% \multicolumn{2}{c}{\textbf{MDVAE}}\\
% \includegraphics[width=0.23\textwidth]{Figures/MDVAE_cLogP.png} &
% \includegraphics[width=0.23\textwidth]{Figures/MDVAE_cLogS.png} \\
% Mol weight&Drug-likeness\\
% \includegraphics[width=0.23\textwidth]{Figures/MDVAE_Molweight.png} &
% \includegraphics[width=0.23\textwidth]{Figures/MDVAE_Druglikeness.png}\\
% \end{tabular}
% \caption{Comparison of the distribution of cLogP, cLogS, Molecular Weight, and Drug-likeness (from left to right) in the generated versus the training dataset for CGVAE (top row), DVAE (middle row), and MDVAE (bottom row). Results are better seen in color.}
% \label{fig:mol_prop}
% \end{figure}

\begin{figure}[htbp]
\centering
\begin{tabular}{cc}    
\multicolumn{2}{c}{\textbf{CGVAE}}\\
clogP&clogS\\
\includegraphics[width=0.23\textwidth]{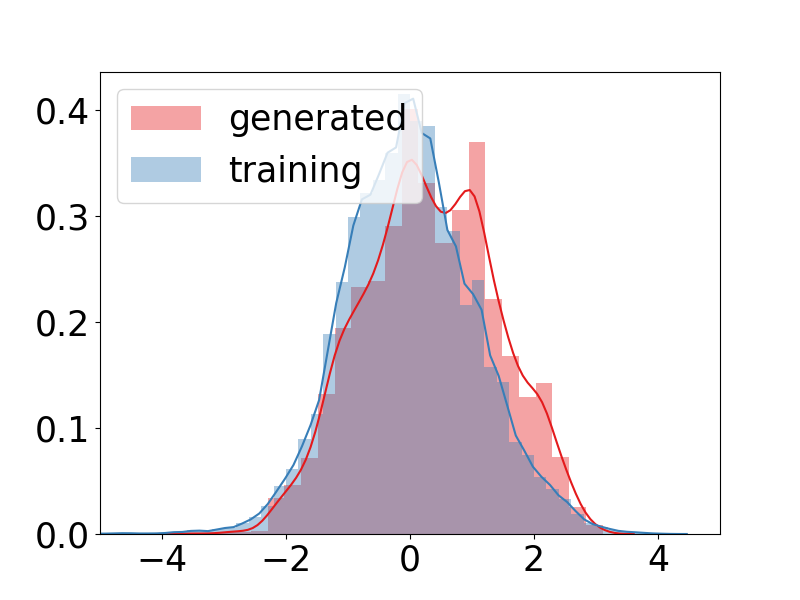} &
\includegraphics[width=0.23\textwidth]{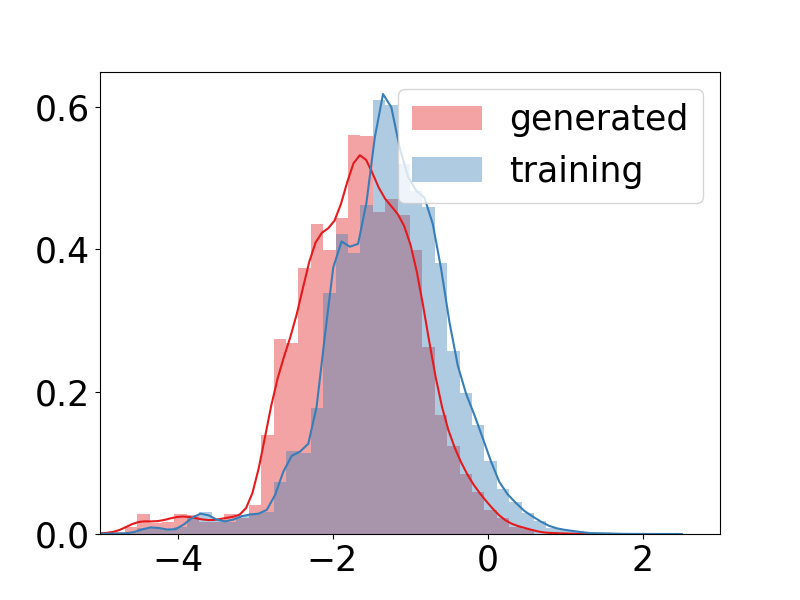} \\
Mol weight&Drug-likeness\\
\includegraphics[width=0.23\textwidth]{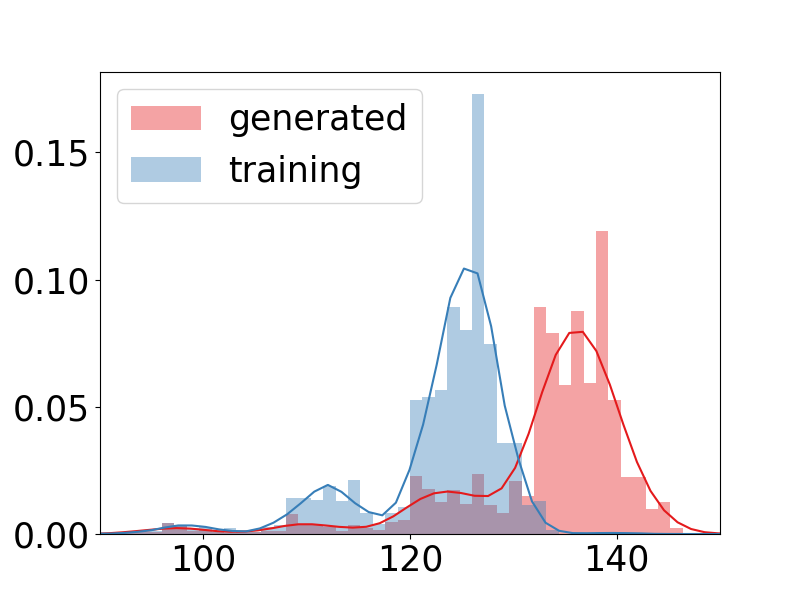} &
\includegraphics[width=0.23\textwidth]{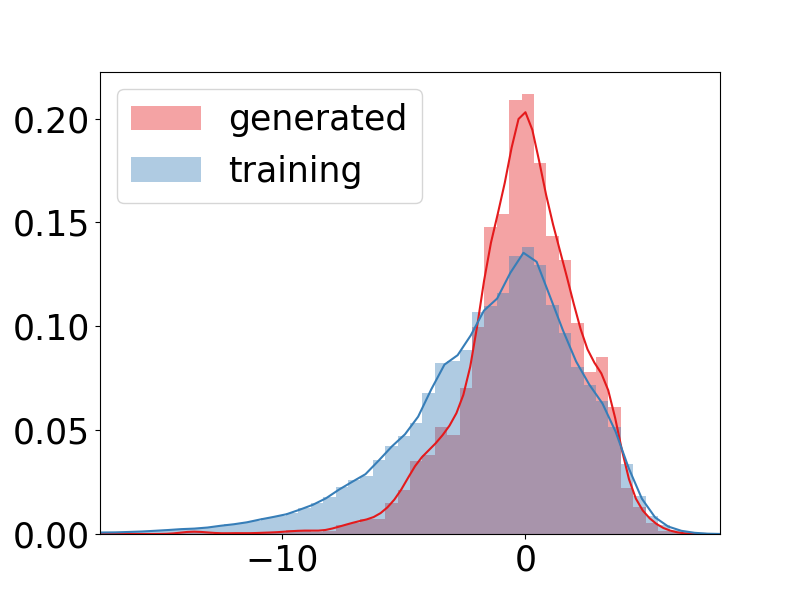} \\
\end{tabular}
\caption{Comparison of the distribution of cLogP, cLogS, Molecular Weight, and Drug-likeness in the generated versus the training dataset for CGVAE). Results are better seen in color.}
\label{fig:mol_prop}
\end{figure}

\begin{figure}[htbp]
\centering
\begin{tabular}{cc}    
\multicolumn{2}{c}{\textbf{CGVAE}}\\
\includegraphics[width=0.23\textwidth]{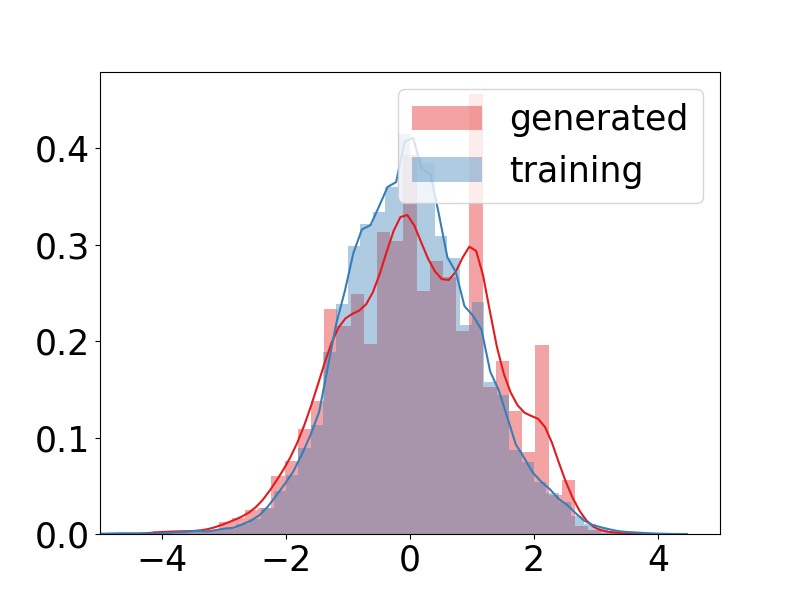} &
\includegraphics[width=0.23\textwidth]{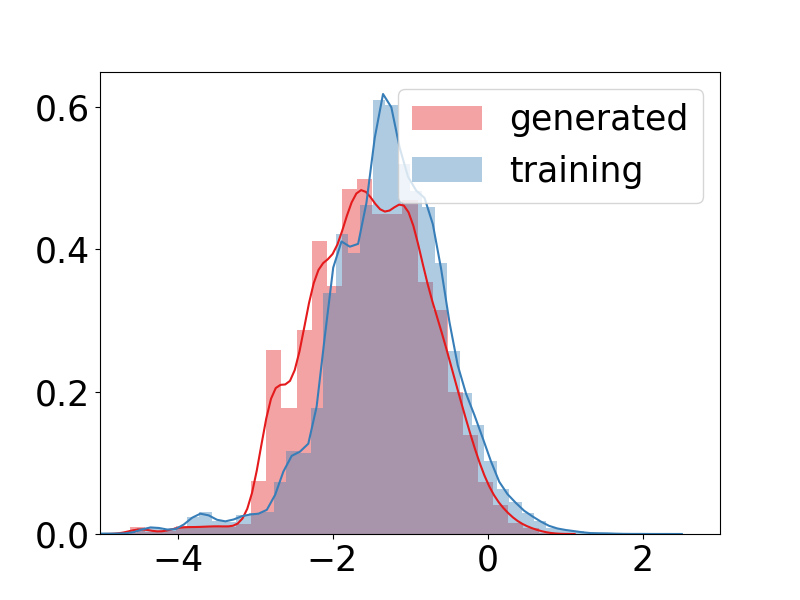} \\
Mol weight&Drug-likeness\\
\includegraphics[width=0.23\textwidth]{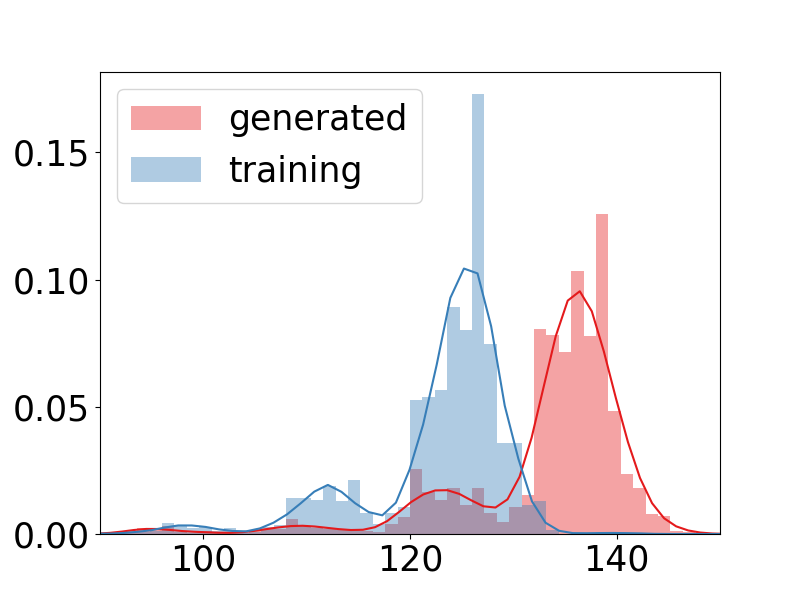} &
\includegraphics[width=0.23\textwidth]{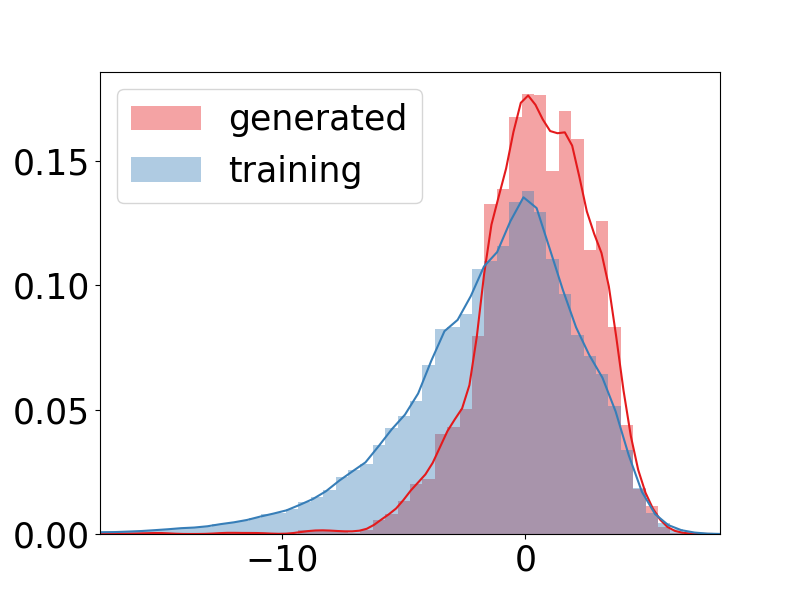}\\
\multicolumn{2}{c}{\textbf{MDVAE}}\\
\includegraphics[width=0.23\textwidth]{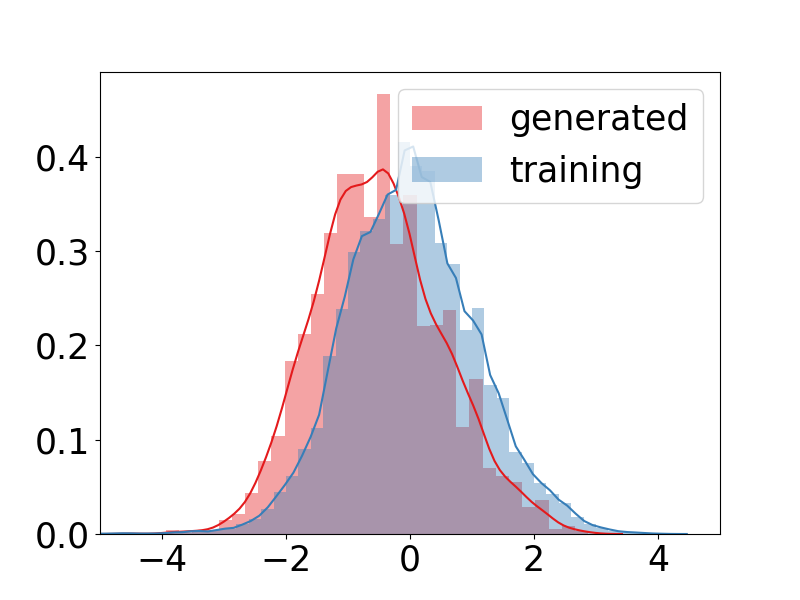} &
\includegraphics[width=0.23\textwidth]{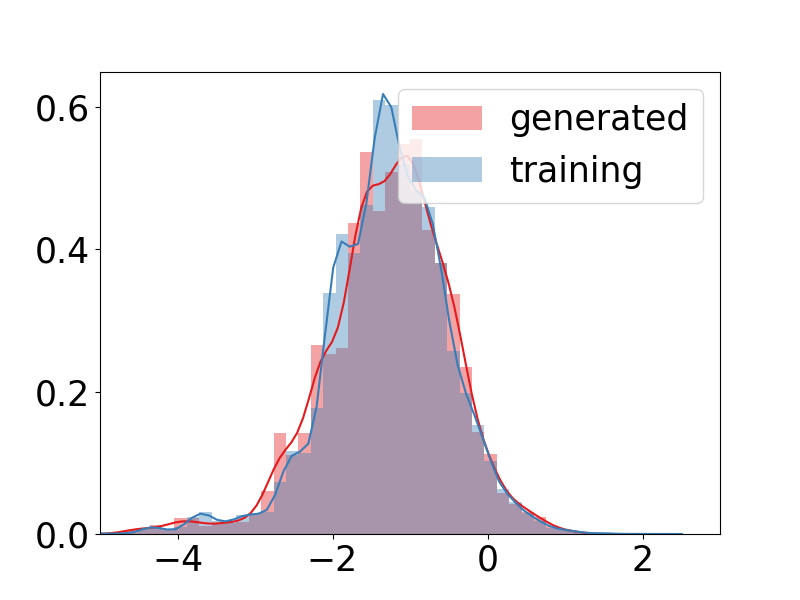} \\
Mol weight&Drug-likeness\\
\includegraphics[width=0.23\textwidth]{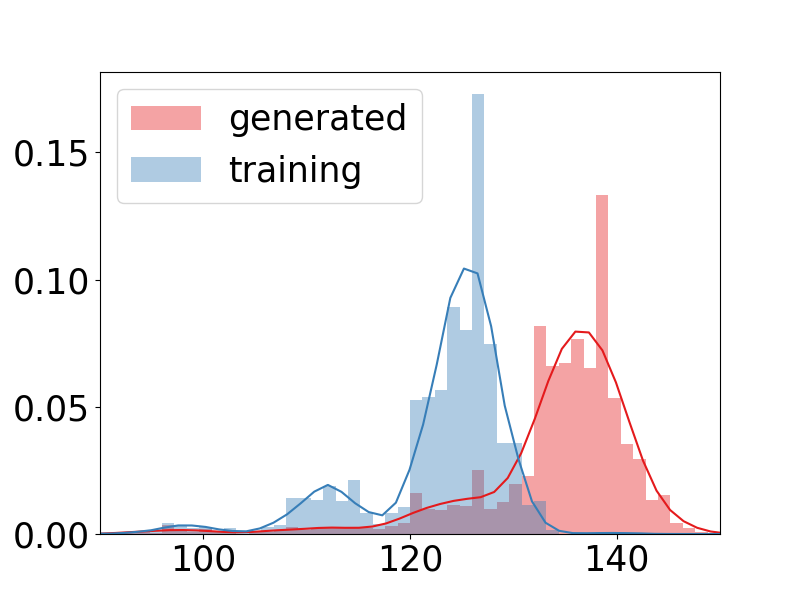} &
\includegraphics[width=0.23\textwidth]{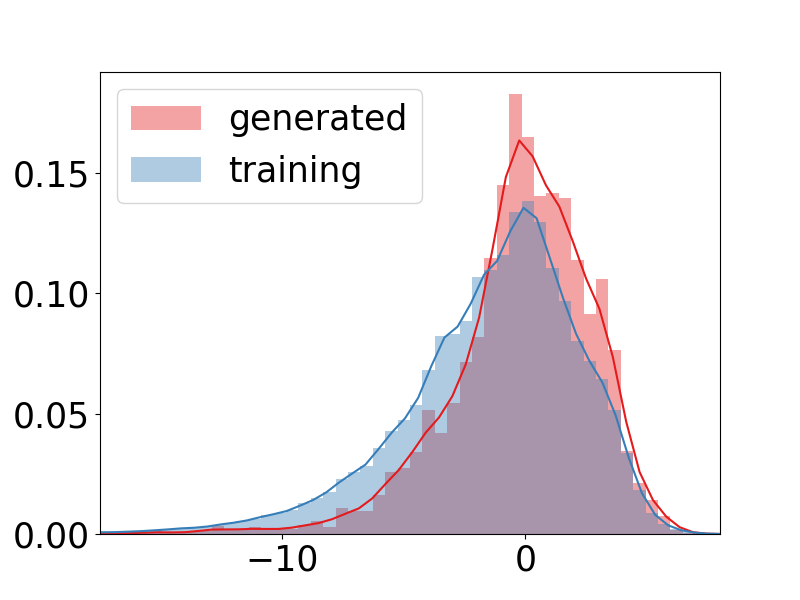}\\
\end{tabular}
\caption{Comparison of the distribution of cLogP, cLogS, Molecular Weight, and Drug-likeness in the generated versus the training dataset for DVAE (top row), and MDVAE (bottom row). Results are better seen in color.}
\label{fig:mol_prop}
\end{figure}

\end{document}